%% file: main.tex
\documentclass[twoside,11pt]{article}
\usepackage{jmlr2e}
\usepackage{subcaption}
\usepackage{amsmath,amssymb,amsfonts}
\usepackage{algorithmic}
\usepackage[algo2e]{algorithm2e}
\usepackage{algorithm}
\usepackage{booktabs}
\usepackage{graphicx}
\usepackage{url}
\usepackage{soul}
\usepackage{textcomp}
\usepackage{flafter}
\usepackage{makecell}

\SetKwComment{Comment}{$\triangleright$\ }{}
\def\BibTeX{{\rm B\kern-.05em{\sc i\kern-.025em b}\kern-.08em
    T\kern-.1667em\lower.7ex\hbox{E}\kern-.125emX}}

\ShortHeadings{SNoRe}{Me\v{z}nar, Lavra\v{c} and \v{S}krlj}
\firstpageno{1}

\begin{document}

\title{SNoRe: Scalable Unsupervised Learning of Symbolic Node Representations}

\author{\name Sebastian Me\v{z}nar \email smeznar@gmail.com \\
       \addr Jo\v{z}ef Stefan Institute\\
       Jamova 39, 1000 Ljubljana, Slovenia
       \AND
       \name Nada Lavra\v{c} \email nada.lavrac@ijs.si  \\
       \addr Jo\v{z}ef Stefan Institute\\
       Jamova 39, 1000 Ljubljana, Slovenia \\
       \addr Jo\v{z}ef Stefan International Postgraduate School \\
       Jamova 39,1000 Ljubljana, Slovenia \\
       \addr University of Nova Gorica \\
        Glavni trg 8, Vipava, Slovenia
       \AND
       \name Bla\v{z} \v{S}krlj \email blaz.skrlj@ijs.si   \\
       \addr Jo\v{z}ef Stefan Institute\\
       Jamova 39, 1000 Ljubljana, Slovenia \\\
       \addr Jo\v{z}ef Stefan International Postgraduate School \\
       Jamova 39,1000 Ljubljana, Slovenia}

\editor{Sebastian Me\v{z}nar}

\maketitle

\begin{abstract}
Learning from complex real-life networks is a lively research area, with recent advances in learning information-rich, low-dimensional network node representations. However, state-of-the-art methods are not necessarily interpretable and are therefore not fully applicable to sensitive settings in biomedical or user profiling tasks, where explicit bias detection is highly relevant. The proposed SNoRe (Symbolic Node Representations) algorithm is capable of learning symbolic, human-understandable representations of  individual network nodes, based on the similarity of neighborhood hashes which serve as features.
SNoRe's interpretable features are suitable for direct explanation of individual predictions, which we demonstrate by coupling it with the widely used instance explanation tool SHAP to obtain nomograms representing the relevance of individual features for a given classification. To our knowledge, this is one of the first such attempts in a structural node embedding setting. In the experimental evaluation on eleven real-life datasets, SNoRe proved to be competitive to strong baselines, such as variational graph autoencoders, node2vec and LINE. The vectorized implementation of SNoRe scales to large networks, making it suitable for contemporary network learning and analysis tasks.
\end{abstract}


\section{Introduction}
\label{sec:introduction}
Networks can be used to model numerous real-world systems, spanning from biological protein interaction networks to social and transportation networks~(\cite{lu2011link,costa2011analyzing}). By representing a real-life system as a network, it is possible to study network properties, such as the key network nodes, why they are relevant, how sets of nodes group together and how network nodes are classified~(\cite{cai2018comprehensive,bhagat2011node}). The latter task is the focus of this research. By using networks to represent real-life systems, we can further explore \emph{interactions between instances} instead of conventional approaches that assume instance independence.

The problem of node classification has been already considered in the 1990s~(\cite{snets}). However, it was popularized only in the recent years due to the increase in the available computational power. A well-known method capable of node classification is label propagation~(\cite{zhu2002learning}), an algorithm that asynchronously assigns labels to neighboring nodes, eventually reaching an equilibrium state that corresponds to the final classification. Albeit efficient, label propagation and similar approaches operate in a relatively na\"ive manner, not accounting for the rich structure of a given network that spans \emph{beyond simple neighborhoods}. To mitigate this issue, novel representation learning methods emerged, offering efficient ways of constructing real-valued vector representations of individual nodes, suitable for down-stream learning such as classification.

Contemporary structural node representation algorithms are mostly concerned with the down-stream performance, with insufficient focus on the \emph{interpretability}, which is of utmost importance when the user tries to understand \emph{why} the system decided to classify a given instance the way it did. To mitigate this issue, we developed SNoRe, an algorithm that compares node neighborhoods and is capable of learning interpretable feature sets through symbolic expressions, describing a given node, which can be used to obtain \emph{explanations} of individual predictions, being an improvement over state-of-the-art low-dimensional, black-box node representations.
The contributions of this work are summarized as follows:
\begin{itemize}
    \item We propose SNoRe, an efficient algorithm capable of learning symbolic representations of nodes by accounting for global network topology.
    \item Theoretical and empirical comparisons with state-of-the-art indicate competitive performance, whilst offering the interpretability of individual predictions, explained by the contributions of the neighboring nodes.
    \item We show that SNoRe scales to real-life networks with tens of thousands of nodes, and does not require dedicated hardware for effective performance.
    \item SNoRe is implemented as a simple-to-use Python library, transpiled to lower-level code via the Numba framework~(\cite{lam2015numba}) for maximum efficiency. The implementation also features a highly efficient sparse implementation of the Hub Promoted Index (HPI).
\end{itemize}

\section{Related work}
\label{sec:related}

This section presents the state-of-the-art methods capable of solving the node classification task along with their properties. Note that there are two main settings for learning from networks, referred to as  \emph{transductive} and \emph{inductive} learning. In the \emph{transductive} learning setting, node classification is performed within the same network, where part of the network is initially labeled, whereas the remaining part is not. The task addresses the issue of \emph{extrapolating} the information from the known part of the network to the unknown (unlabeled) part. Common examples of this task include gene function prediction and social network-based tasks, such as user profiling. On the other hand, in the \emph{inductive} learning setting, independent networks are fed as input and are also classified on the network level. The focus of this work is on \emph{transductive} learning.

The types of learning algorithms can further be split based on the information they are capable to exploit during learning. An algorithm can perform solely by exploiting the network structure, or can also incorporate potentially interesting features assigned to nodes or edges. The focus of this work is on \emph{structural classification} with no assigned features.

\subsection{Structural node embedding}
\label{sec:strucnn}
The notion of \emph{structural node embedding} corresponds to the process of learning a given node's latent representation (most commonly real-valued), based on its neighborhood within a given network. The first branch of methods was inspired by the widely known word2vec algorithm~(\cite{NIPS2013_5021}):  DeepWalk~(\cite{Perozzi2014deepwalk}) was one of the first node representation learners, and remains state-of-the-art to this date. DeepWalk creates a network representation by using sequences of nodes representing random walks as input sentences for the word2vec algorithm. Random walks created in a depth-first search manner intuitively map nodes with similar second-order proximity close together.

Following similar ideas, methods such as node2vec, struc2vec, LINE, PTE, NetMF and others emerged, each considering additional network properties (e.g., network-topological properties) during representation learning. Algorithm node2vec~(\cite{grover2016node2vec}) uses hyper-parameters $p$ and $q$ to guide random walks. Parameter $p$ dictates the return probability whereas $q$ dictates the probability of exploration away from the previous node. If $p$ and $q$ are set to $1$ we get the special case where the node2vec algorithm is equivalent to DeepWalk. 

LINE~(\cite{tang2015line}) derives an objective function for first and second-order proximity that is computationally intensive and thus not scalable. The algorithm is then made scalable with the adoption of negative sampling. Function parameters of the classification model are optimized with asynchronous stochastic gradient descent.

NetMF~(\cite{qiu2018network}) is presented along with the theoretical analysis of DeepWalk~(\cite{Perozzi2014deepwalk}), node2vec~(\cite{grover2016node2vec}), LINE~(\cite{tang2015line}) and PTE~(\cite{Tang2015PTEPT}), showing that all the aforementioned methods approximate matrix factorization and that the close forms of these matrices are intrinsically connected to the graph Laplacian. NetMF factorizes these closed form matrices, potentially offering consistent improvement in performance over the methods mentioned above.

Personalized Page Rank with Shrinking (PPRS) was introduced as a part of the HINMINE methodology (\cite{kralj2017hinmine}). This algorithm creates vectors representing personalized node ranks by using the power iteration. Such vectors can be used directly for learning purposes, or further compressed by an autoencoder~(\cite{skrlj2019deep}), offering small compact representations trained in an end-to-end manner.

Similarly to these embedding algorithms, our proposed approach uses information about a given node's neighborhood to create a representation in unsupervised manner. Instead of creating a dense, latent embedding, our algorithm returns a sparse embedding where features represent nodes, which makes the result easily interpretable.

\subsection{Graph neural networks}
\label{sec:gnn}
Since networks as such are not bound to a given coordinate system,  direct input of e.g., adjacency matrices into neural networks proves to be problematic. As a result, in parallel with the aforementioned structural node embedding  methods, which are useful for representation learning in domains with a well structured spatial structure (such as images), the area of \emph{graph neural networks} (GNNs) (\cite{zhang2020deepsurvey,wu2020gnns,xu2018jumping, Bojchevski2019IsPA, klicpera_predict_2019}) emerged, conceived to tackle the problem of learning from unstructured domains.

Graph Neural Networks operate by passing feature information from the considered node's neighbors towards the node itself. During this process, the latent representation (embedding) of the node is obtained. The final representation is, in most cases, a result of gradient descent-based optimization. These algorithms are mostly divided into three subgroups: graph recurrent neural networks, graph convolutional neural networks and graph autoencoders. Graph recurrent neural networks try to capture and learn recursive and sequential patterns by taking advantage of recurrent neural networks. Graph convolutional neural networks learn local and global patterns trough designed convolution and readout functions. 

Graph convolutional neural networks are divided into spectral and spatial based algorithms, based on how they define convolution. Spectral based algorithms define graph convolution using filters from graph signal processing, whereas the convolution in spatial algorithms relays on information propagation. Graph autoencoders are often used for unsupervised representation learning by assuming that the networks have low-rank structures that are potentially nonlinear~(\cite{zhang2020deepsurvey}).

Graph convolutional networks (GCNs)~(\cite{kipf2016semi}) are among the most influential works in graph-based deep learning since CNNs bridge the gap between spectral and spatial based graph convolutional neural networks. The GCN algorithm simplifies filtering by only focusing on first-order neighbors. Since the number of neighbors can vary, GraphSAGE~(\cite{hamilton2017graphsage}) samples a fixed amount of neighbors and aggregates them. The Graph attention network (GAT)~(\cite{velickovic2018graph}) further improves both previously mentioned approaches by introducing the attention mechanism. Attention mechanism allows the neural network to learn how much each neighbor contributes instead of assuming that all neighbors contribute the same amount (like in GraphSAGE) or that this amount is predetermined (like in GCN). Another interesting graph convolutional neural network is the Graph Isomorphism Network (GIN)~(\cite{xu2018powerful}) that presents a readout function that uses summation and a multi-layer perceptron to provably achieve the maximum discriminative power.

A popular graph autoencoding algorithm is the Variational graph autoencoder (VGAE)~(\cite{kipf2016variational}) that uses latent variables to create a representation for undirected networks. The algorithm encodes the network into mean and variance matrices and decodes them with the dot product. The parameters of the model are learned by minimizing the variational lower bound.

While Graph Neural Networks represent the state-of-the-art in node classification, they differ significantly from our approach as they usually use features that are not calculated from the network and classify nodes in an end-to-end fashion.

\section{The SN{o}R{e} algorithm}
In this section we first define some essential components and present the key ideas of the SNoRe algorithm (Section~\ref{sec:definitions}). The algorithm is divided into four steps: random walk generation (Section~\ref{sec:randomgen}), random walk hashing (Section~\ref{walkhash}), feature selection (Section~\ref{sec:featselection}), and similarity calculation (Section~\ref{sec:similaritycalc}). For each step, we present its description and its implementation. We also propose an extension of the algorithm that chooses the number of features based on the embedding size (Section~\ref{sec:dim}), show an overview of the algorithm (Section~\ref{sec:overview}) and present its theoretical properties (Section~\ref{sec:properties}).

\subsection{Definitions and key ideas}
\label{sec:definitions}
Let us first define the key terms used throughout this paper.

\textbf{Definition 1 (Network).} \textit{ A network is a tuple $G = (N, E)$, where $N$ represents the set of  nodes and $E$ represents the set of edges. An edge can be represented as an ordered pair (e.g., $(n_1, n_2) \in N \times N $)---in this case the network is directed. Alternatively, an edge can be represented as a subset of size 2 (e.g., $ \{n_1, n_2\} \subseteq N$)---in this case the network is undirected.}

For generality, we will use directed networks since we can also represent the undirected ones using the same formalism. We define a walk and a random walk as follows.

\textbf{Definition 2 (Walk).} \textit{ A walk of length $k$ in a directed network is any sequence of $k$ nodes $n_1, n_2, ..., n_k \in N$, so that each pair of consecutive nodes $n_i$ and  $n_{i+1}$ has a connection $(n_i, n_{i+1}) \in E$.}

\textbf{Definition 3 (Random walk).} \textit{A random walk is a walk generated in such way that at step $i$, node $n_{i+1} \in \{a, (n_i, a) \in E\}$ is chosen with some probability.}

The result of our algorithm is symbolic node embedding of a network defined as:

\textbf{Definition 4 (Symbolic node embedding).} \textit{ Symbolic node embedding of network $G = (N, E)$ is a matrix $\boldsymbol{M} \in \mathbb{R}^{|N| \times d} $, where $d$ is the dimension of the embedding. Such embedding is considered symbolic, when each column represents a symbolic expression, which---when evaluated against a given node's neighborhood information---returns an integer number representing a given node.}
This definition uses the term symbolic expression to describe the structure of data that can be easily interpreted by a human (i.e. the similarity between neighborhoods of nodes $i$ and $j$).

Note that the above defined type of symbolic node embedding can also be referred to as propositionalization (see a recent review~(\cite{Lavrac2020}) for more details).

\begin{figure*}[htb!]
  \centering
  \includegraphics[width =.75 \linewidth] {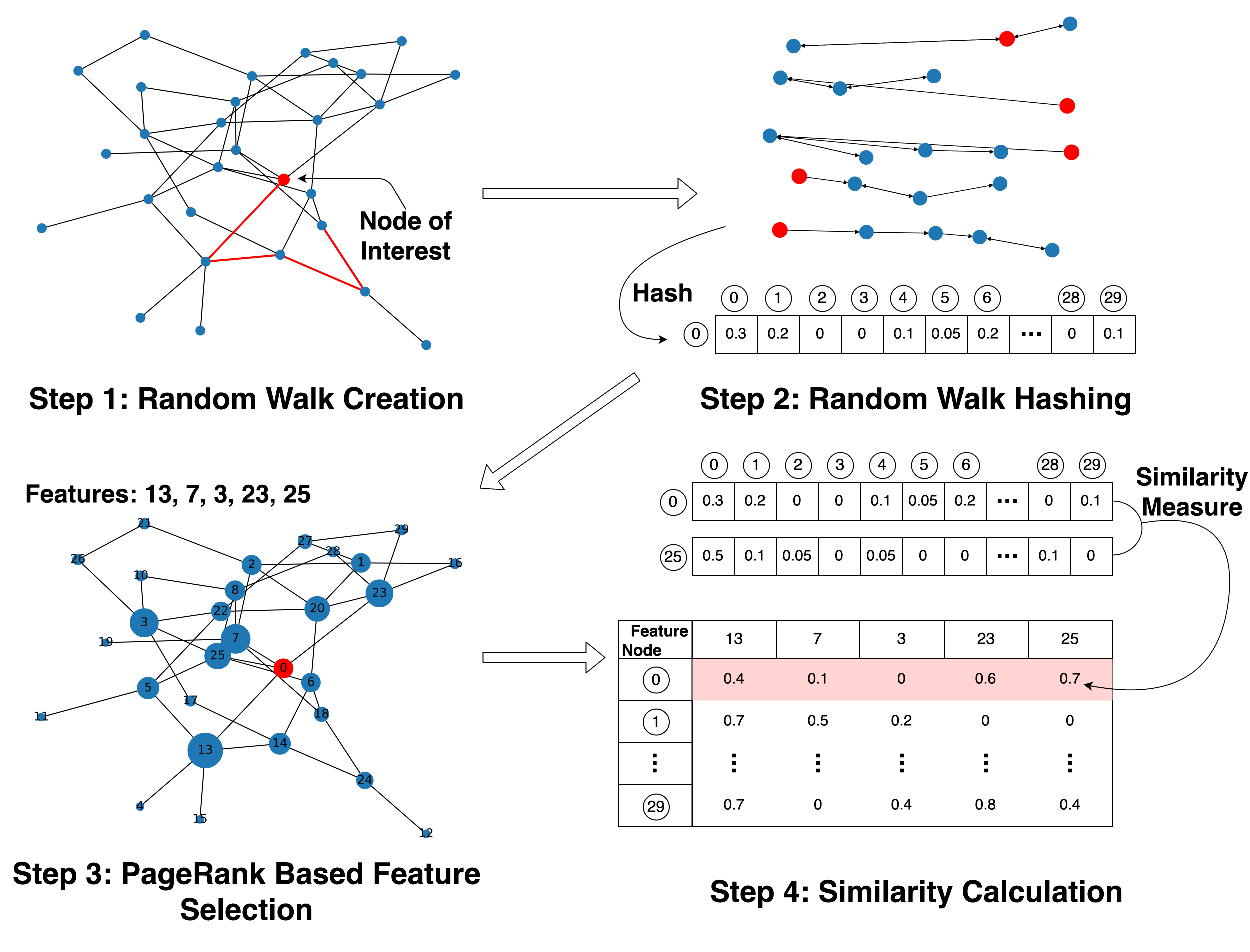}
  \caption{SNoRe key idea overview. Step 1 generates random walks that are then hashed in Step 2. These hashes are represented as sparse vectors and used to calculate the similarity between two node neighborhoods in Step 4, where the similarity is calculated between all nodes and the nodes that are chosen as features in Step 3 based on their PageRank score.}
  \label{fig:overview}
\end{figure*}

We use the above definitions to outline the proposed SNoRe algorithm, illustrated in Fig.~\ref{fig:overview}. In the figure, we highlight a node and mark it red to present how an arbitrary node in the network gets embedded. The first step generates random walks, marked as a collection of red edges in Fig.~\ref{fig:overview}. We then aggregate walks starting at the red node into a vector of node occurrences in step 2. Step 3 then selects the features based on weights assigned by the PageRank algorithm~(\cite{ilprints422}). In the final step, we generate the embedding of any given node by calculating cosine similarity between the hash values of nodes selected as features and the red (considered) node.

\subsection{Random walk generation}
\label{sec:randomgen}
Sampling the neighborhood of a given node can give us information about the network structure and the connectivity patterns in its vicinity. We can sample the neighborhood using short random walks. These offer many advantages such as ease and parallelization of computation, bound for the distance of the farthest node and ease of representation.

The first step of the algorithm generates random walks and represents them with a data structure such as a list of visited nodes. We use the random walk generation scheme (and vectorized implementation) presented in~(\cite{skrlj2019sge}). Let $\boldsymbol{w} \in \mathbb{R}^s, \left\lVert \boldsymbol{w} \right\rVert_1 = 1$ be the distribution vector, where $s$ is the maximum length of the walk and $\boldsymbol{w}_i$ denotes the probability that the walk is of length $i$. We sample random walk length $i$ from $\boldsymbol{w}$ and create a random walk of length $i$ using Algorithm~\ref{alg1}. In line $4$ of the algorithm we append the current node $c$ (together with some information) to the walk representation structure. Function $\textrm{neighbor}$ in line $5$ returns a neighbor of the given node (randomly). This algorithm is repeated $\textrm{nw}$ times for each node, giving us $\textrm{nw}$ random walks per node.

\begin{algorithm}[h]
\caption{Classical walk}
\label{alg1}
\SetKwInOut{Input}{Input}\SetKwInOut{Output}{Output}
\Input{Starting node $n_i$, Walk length $\textrm{wl}$}
\Output{Random walk structure $\textrm{Ws}$}
\begin{algorithmic}[1]
    \STATE $c \gets n_i$
    \STATE $\textrm{Ws} \gets \emptyset$
    \FOR{$i=0$ \TO $\textrm{wl}$ } 
        \STATE $\textrm{Ws} \gets \textrm{Ws} \cup c$
        \STATE $c \gets \textrm{neighbor}(c)$
    \ENDFOR
    \RETURN $\textrm{Ws}$\;
\end{algorithmic}
\end{algorithm}

In our implementation, we represented a random walk with a list of tuples denoting the node and step $(n_i, s_i) \in N \times \{j, j=0, ..., s\}$. The final random walk structure consists of concatenated random walk lists $l_i$ for each node separately.

The time and space complexity of computing the random walk structure is $\mathcal{O}(|N| \cdot \textrm{nw} \cdot \overline{s})$, where $\overline{s}$ represents the mean length of the walk. We get this time complexity because for each node we create $\textrm{nw}$ walks that make $\overline{s}$ steps on average. Since only the random walk hashing step uses this representation of random walks, the space complexity can drop to a constant if we merge the first two steps by incrementally calculating the hash value after each walk. This way the walks \emph{do not have to be stored}.

\subsection{Random walk hashing}
\label{walkhash}
We represent the neighborhood of node $n_i$ numerically by hashing random walks starting in $n_i$. The hashing function can incorporate different sources of information about the network to make a vector, $\boldsymbol{h} \in \mathbb{R}^{\textrm{dh}}$, where $\textrm{dh}$ is the dimension of hashing function output. Some examples of this can include: occurrences of nodes normalized, the number of nodes with some degree normalized or occurrences of the communities normalized. We will denote the hash value (vector) for the $i$-th node as $\boldsymbol{h}_i$.

Our implementation uses only neighborhood-level information about the network, i.e. how often a node appears in random walks that start at node $n_i$. We also use threshold $\epsilon$ as the lower bound for occurrences. Any node that occurs less then $length(l_i)\cdot\epsilon$ times is not included in $\boldsymbol{h}_i$. The final hash is a normalized sparse row vector, where values represent how frequently an included node was encountered during a random walk.

\subsection{Feature selection}
\label{sec:featselection}
Features of the node embedding created by our algorithm are symbolic expressions that can be easily interpreted. We use a subset of nodes as our features to satisfy this goal. The feature values represent the similarity between the neighborhoods of a given node and the node that represents the feature. We will use $\textrm{feature-map}:\mathbb{N}\to N$ as the function mapping feature index to the corresponding node. 

Feature selection can be done in a supervised or unsupervised manner~(\cite{saeys2007feature}). We focus on unsupervised feature selection so that the whole algorithm can remain unsupervised.

In feature selection, we want to select nodes that are important for the network structure. We assign a score to each node using the PageRank algorithm~(\cite{ilprints422}), then sort them based on this score in the descending order, and select top $d$ nodes as our features.

The PageRank algorithm computes a probability distribution $\boldsymbol{pr} \in \mathbb{R}^{|N|}, \left\lVert \boldsymbol{pr} \right\rVert_1=1$, where $\boldsymbol{pr}_i$ approximates the probability of a random walker being at  node $i$. When $\boldsymbol{pr}_i$ is high, node $i$ is more likely to be visited and therefore it is likely more important for the structure of the network. Let $\boldsymbol{r} \in \mathbb{R}^{|N|}$ represent a vector of PageRank values for each node. Let $d_{j}$ represent the degree of the $j$-th node. If the adjacency matrix of the considered network ($\boldsymbol{A}$) is normalized as follows:
\begin{equation*}
    \boldsymbol{C}_{ij} = \begin{cases}
    \frac{1}{d_{j}}; \boldsymbol{A}_{ij} \neq 0 \\
    0; otherwise
    \end{cases}
\end{equation*}
\noindent the computation of PageRank can be formulated as an eigenvalue problem:
\begin{equation*}
    \boldsymbol{r} = \boldsymbol{C}\boldsymbol{r}.
\end{equation*}
For larger networks, the \emph{power iteration} is used to approximate the final solution. This procedure first initializes $\boldsymbol{r} = [\frac{1}{|N|},\frac{1}{|N|},\dots,\frac{1}{|N|}]^T$ (i.e. a discrete uniform distribution), and iterates by computing:
\begin{equation*}
    \boldsymbol{r}_{k+1} = \boldsymbol{C} \boldsymbol{r}_k,
\end{equation*}
\noindent until the difference between $\boldsymbol{r}_k$ and $\boldsymbol{r}_{k+1}$ is smaller than some predetermined threshold $\mu$. The final $\boldsymbol{r}$ represents the final collection of PageRank values considered in this work. Note that in practice, about 10--50 iterations are needed for convergence, making this method highly scalable.

We chose PageRank as the scoring function used in feature selection because it is fast, unsupervised and gives a good approximation for node importance. This choice, however, is not very important since many times, choosing nodes randomly gives us only slightly worse results. This is especially true for sparse networks coupled with the extension of SNoRe we present in Section~\ref{sec:dim} since most if not all nodes are chosen. This extension uses feature ranking to estimate $d$ such that the embedding we get is equivalent in size to a chosen dense embedding.

PageRank has a parameter, alpha, also known as the damping factor. In our work, we used the value 0.85 as the parameter since it is the default value in the NetworkX~(\cite{SciPyProceedings_11}) implementation we used.

\subsection{Similarity calculation}
\label{sec:similaritycalc}
The proposed SNoRe algorithm creates a symbolic node embedding matrix $\boldsymbol{M}$, where row $\boldsymbol{m}_i$ represents the similarity of the $i$-th node to the nodes chosen as features (pivot nodes). This similarity is calculated in the final step from hash values $\boldsymbol{h}_i$ generated in the random walk hashing step. We compare the hash value $\boldsymbol{h}_i$ of the $i$-th node to the hash value $\boldsymbol{h}_{\textrm{feature-map}(j)}$ of the $j$-th pivot node.

The cosine similarity metric is defined such that it represents the cosine angle between two non-zero vectors: \begin{align*}
cos\_sim(\boldsymbol{a}, \boldsymbol{b}) = \frac{\sum_{i=1}^{|N|}\boldsymbol{a}_i\cdot \boldsymbol{b}_i} {\sqrt{ \sum_{i=1}^{|N|} \boldsymbol{a}_{i}^2} \sqrt{\sum_{i=1}^{|N|} \boldsymbol{b}_{i}^2}},
\end{align*}
where $\boldsymbol{a}$ and $\boldsymbol{b}$ represent the two vectors\footnote{We use scikit-learn implementation~(\cite{scikit-learn}) for efficient cosine similarity calculation between sparse vectors.}. The similarity score between two vectors without common features is $0$, and between two vectors with the same angle is $1$. This makes the similarity between two vectors easily interpretable. Further, since the score can be $0$, this metric works well with sparse representations. Because of these properties, we use cosine similarity as our main similarity calculation metric.
In Section~\ref{sec:distances} we further demonstrate the advantage of cosine similarity and show how different distance measures compare against it.

\subsection{Estimating the representation dimension}
\label{sec:dim}
One of the key features of SNoRe is its ability to construct \emph{sparse} representations of individual nodes. Compared to e.g., DeepWalk and similar methods, where the dimension is predetermined, SNoRe exploits the following theoretical insight to construct a high dimensional representation with \emph{the same (or lower)} memory footprint than the comparative methods.
As the dimensions in SNoRe can be computed independently (walks w.r.t. individual nodes are independent), this feature offers an iterative expansion of the representation until a sufficient number of e.g., floating-point values is obtained.

The following example demonstrates the mentioned functionality. Consider a situation where SNoRe is to be compared against a dense representation learning algorithm, which learns $d$ dimensional representations of nodes. Assuming $|N|$ instances, the total space required to store the representation can be denoted with $\tau = |N| \cdot d$ (floating-point values). The SNoRe algorithm constructs the representation requiring the same (or less) space in the following manner.
We follow the first three steps of the algorithm to create hash values and the list of nodes sorted in the descending order by their PageRank score. We then add features incrementally calculating the similarity between the added feature and all nodes. After each calculation, we subtract the number of nonzero values for the feature from $\tau$ and return the created embedding when $\tau$ drops below zero.

During testing, we realized that the quality of the embedding is not affected much by small changes in the similarity score and that sometimes digitizing it helps classification (showcased in  Appendix~\ref{sec:app-dig}). For this reason, we divide the interval $[0, 1]$ into $b$ sub-intervals, where $\text{sub-interval}_i = [\frac{i}{b}, \frac{i+1}{b})$, and use them to discretize the similarity score between two hashes. We replace the similarity score $\boldsymbol{M}_{i,j}$ with $\frac{\textrm{idx}}{b}$, where $\textrm{idx}$ denotes the index of sub-interval containing $\boldsymbol{M}_{i,j}$. This allows us to store values using fewer bits and consequently create the embedding with more features that takes up the same amount of space.

The automatic \emph{sparse} representation construction is outlined in lines 12--25 of Algorithm~\ref{alg2}. This extension is presented as SNoRe with Size Dependent Features (SNoRe (SDF)) in Section~\ref{sec:results}. The behaviour of SNoRe and SNoRe (SDF) only differs in the number of nodes used in the final embedding, so we use the terms interchangeably in the rest of the paper. 

\subsection{SNoRe overview}
\label{sec:overview}
The pseudocode of SNoRe (with the final step of SNoRe (SDF)) is presented in Algorithm~\ref{alg2}. Function $\textrm{SAMPLE}$ takes a distribution vector described in Section~\ref{sec:randomgen} as the input and returns an integer representing walk length sampled from it. Function $\textrm{WALK}$ returns a structure that represents a random walk and takes as arguments the starting node and the walk length. Function $\textrm{HASH}$ returns the hash value of the inputted walks. Function $\textrm{PAGE\_RANK}$ returns a sorted list of nodes based on their PageRank scores. Function $\textrm{SIM}$ returns a number between $0$ and $1$ that represents the similarity between two hashes given as input (distance between the obtained walk distributions).

\begin{algorithm}
\caption{SNoRe (SDF)}
\label{alg2}
\SetAlgoLined
\SetKwInOut{Input}{Input}\SetKwInOut{Output}{Output}
\Input{Network $G=(N, E)$, Length distribution $\boldsymbol{w}$, Maximum size $\tau$, Number of walks $\textrm{num\_walks}$}
\Output{Symbolic node embedding matrix $\boldsymbol{M}$} 
\begin{algorithmic}[1]
    \STATE $\textrm{walks} \gets \emptyset$ \Comment*[r]{Random walk generation.}
    \FOR{$i=1$ \TO $|N|$ } 
        \FOR{$j=1$ \TO $\textrm{num\_walks}$}
            \STATE $\textrm{walks}_i \gets \textrm{walks}_i \cup \textrm{WALK}(N_i, \textrm{SAMPLE}(w))$
        \ENDFOR
    \ENDFOR   
    \STATE $\boldsymbol{h} \gets \emptyset$ \Comment*[r]{Random walk hashing.}
    \FOR{$i=1$ \TO $|N|$ } 
        \STATE $\boldsymbol{h}_i \gets \textrm{HASH}(\textrm{walks}_i)$
    \ENDFOR   
    \STATE $\textrm{feature-map} \gets \textrm{PAGE\_RANK}(G)$ \Comment*[l]{Unsupervised feature ranking.}
    \STATE $\boldsymbol{M} \gets [0]^{|N| \times |N|}$
    \STATE $l \gets 0$ \Comment*[r]{Embedding generation.}
    \WHILE{$\tau \geq 0 \And l < |N|$}
        \STATE $l \gets l + 1$
        \STATE $\textrm{num} \gets 0$
        \FOR{$j=0$ \TO $|N|$ } 
            \STATE $s \gets \textrm{SIM}(\boldsymbol{h}_i,\boldsymbol{h}_{\textrm{feature-map}(j)})$ \Comment*[r]{Similarity.}
            \STATE $\boldsymbol{M}_{i,j} \gets \frac{\textrm{round}(s)}{b}$
            \IF{$\boldsymbol{M}_{i,j} \neq 0$}
                \STATE $\textrm{num} \gets \textrm{num} + 1$
            \ENDIF
        \ENDFOR
        \STATE $\tau \gets \tau - \textrm{num}$
    \ENDWHILE
    \RETURN $\boldsymbol{M} \in \mathbb{R}^{|N| \times l}$
\end{algorithmic}
\end{algorithm}

Lines $1$--$6$ show the random walk generation step. The outer loop iterates over nodes and the inner loop over random walks for each node. In line $4$ the generated random walk is transformed into a suitable representation and appended to the ones already generated. In the implementation, we use memoization to sample walk lengths once and use them for all nodes instead of sampling the length of each walk independently. The generated $\textrm{walks}$ are used in the random walk hashing step that is outlined in lines $7$--$10$.

Hash values (vectors) are generated in the loop shown in lines $8$--$10$. Since hashes are independent between nodes we parallelized this step in the implementation.

The version of the algorithm described in pseudocode also estimates the representation dimension as shown in Section~\ref{sec:dim}. This is done in lines $11$--$25$. Line $11$ calculates the PageRank score of nodes and sorts them. The embedding is iteratively calculated in lines $14$--$25$, adding one feature in each pass until $\tau < 0$ or we run out of features that can be added. We can see that the estimation also uses the similarity calculation step denoted in lines $17$--$23$.
The algorithm finishes in line $26$ where it returns the embedding of size $|N| \times l$, with $\leq \tau = |N| \cdot d$ floating point values.\footnote{Stored using 16-bit NumPy~(\cite{van2011numpy,oliphant2006guide}) type float16.}

\subsection{Theoretical properties}
\label{sec:properties}
For an algorithm to be useful it has to have time and space complexities that are not too resource-intensive. Using the definitions from the previous sections and the understanding of how the algorithm works we next derive the time and space complexities of SNoRe.

\subsubsection{Time complexity}
\label{sec:prop-time}
To present the time complexity we describe how each step of the algorithm behaves and sum the gathered complexities. We simultaneously describe the time complexity of random walk creation and the hashing step, since they can be implemented together efficiently as described in Section~\ref{sec:randomgen}. 

Random walk creation and hashing are computed in $\mathcal{O}(|N|\cdot\textrm{nw}\cdot\overline{s})$, since we need to create $\textrm{nw}$ walks with an average of $\overline{s}$ steps for $|N|$ nodes, whilst assuming that every step takes $\mathcal{O}(1)$ time. Hashing maintains this complexity since each of $|N|\cdot\textrm{nw}$ walks needs $\mathcal{O}(\overline{s})$ time to be hashed.

The time complexity of feature selection depends mostly on the algorithm used for selecting the representative subset of nodes. For feature ranking we used the PageRank algorithm with time complexity $\mathcal{O}(c\cdot |E|)$, when networks are represented with a sparse adjacency matrix. In the aforementioned time complexity $c$ represents the maximum number of iterations. We also need additional $\mathcal{O}(|N|\cdot\log|N|)$ to sort feature scores and gather first $d$ pivot nodes. This can be done more efficiently by only selecting top $d$ pivot nodes, but we rank all nodes for use in the extension. To calculate the time complexity of the last step we focus on the time needed to calculate the similarity between two hash values since this has to be calculated $|N|\cdot d$ times to create the final node embedding matrix. We use sparse implementation of the cosine similarity function with sparse vectors containing at most $\lfloor\frac{1}{\epsilon}\rfloor$ non-zero values. Because of this, we need $\lfloor\frac{1}{\epsilon}\rfloor$ time to compute the similarity between two hashes. Consequently we need $\mathcal{O}(|N| \cdot d \cdot \lfloor\frac{1}{\epsilon}\rfloor)$ to calculate the similarity between each node and each feature.

The algorithm extension that is shown in Section~\ref{sec:dim} only impacts the size of $d$, since other used operations do not contribute significantly to time complexity and can be omitted because of this. Since we use nodes as features $d \leq |N|$ still holds.
Summing the time complexity of all steps we get the following time complexity:
\begin{align*}
&\mathcal{O}(|N|\cdot \textrm{nw}\cdot \overline{s} + c\cdot |E| + |N|\cdot \log|N| + |N|\cdot d \cdot \lfloor\frac{1}{\epsilon}\rfloor) \\&= \mathcal{O}(|N|\cdot (d\cdot \lfloor\frac{1}{\epsilon}\rfloor + \textrm{nw}\cdot \overline{s} + \log|N|) + c\cdot |E|)
\end{align*}

\subsubsection{Space complexity}
The space complexity can be calculated similarly to time complexity by considering the four parts of the algorithm and merging the random walk creation and hashing step. Furthermore, we need $\mathcal{O}(|E|)$ for the sparse adjacency matrix to represent the network. 

We can compute the random walk creation and hashing steps in $\mathcal{O}(\lfloor\frac{|N|}{\epsilon}\rfloor)$ space. Since random walks and hash value calculation can be done for each node independently, we need $\mathcal{O}(\textrm{nw}\cdot\overline{s})$ space for random walk creation and $\mathcal{O}( \lfloor \frac{1}{\epsilon} \rfloor)$ space to store the sparse vector that represents the hash value for this node. This holds because at most $\lfloor \frac{1}{\epsilon} \rfloor$ values can be greater than the threshold $\epsilon$. Since node occurrence is usually not uniform and many nodes occur more frequently than $\epsilon$, the used space is usually smaller than this. We get the space complexity $\mathcal{O}( \lfloor\frac{|N|}{ \epsilon}\rfloor)$ for this two steps by concatenating hash representations of each node. 

The space complexity of the feature selection depends on $d$ and the space complexity of the algorithm used for feature selection. We use PageRank that uses $\mathcal{O}(E)$ space to store a sparse adjacency matrix. We also need $\mathcal{O}(d)$ to store the selected features.

The similarity calculation step creates a (sparse) matrix of size $|N|\cdot d$ where $d \leq |N|$. To calculate the similarity between two hashes we only need constant additional space. If we put the space complexity of all steps together we get the final space complexity:
\begin{align*}
\mathcal{O}(|E| + \frac{|N|}{\epsilon} + |N|\cdot d).
\end{align*}

We further extend the analysis of space complexity with the algorithm extension in Section~\ref{sec:dim} since we generate a sparse matrix that uses less or equal than $\tau = |N|\cdot d$ space, where $d$ is the dimension of a dense embedding.

\section{Datasets and experimental setting}
In this section, we describe the datasets used to evaluate the performance of the proposed embedding algorithm, the experimental setting and the baselines we used to compare the results with.

\subsection{Datasets}
\begin{figure}[t!]
  \centering
  \includegraphics[width = \linewidth]{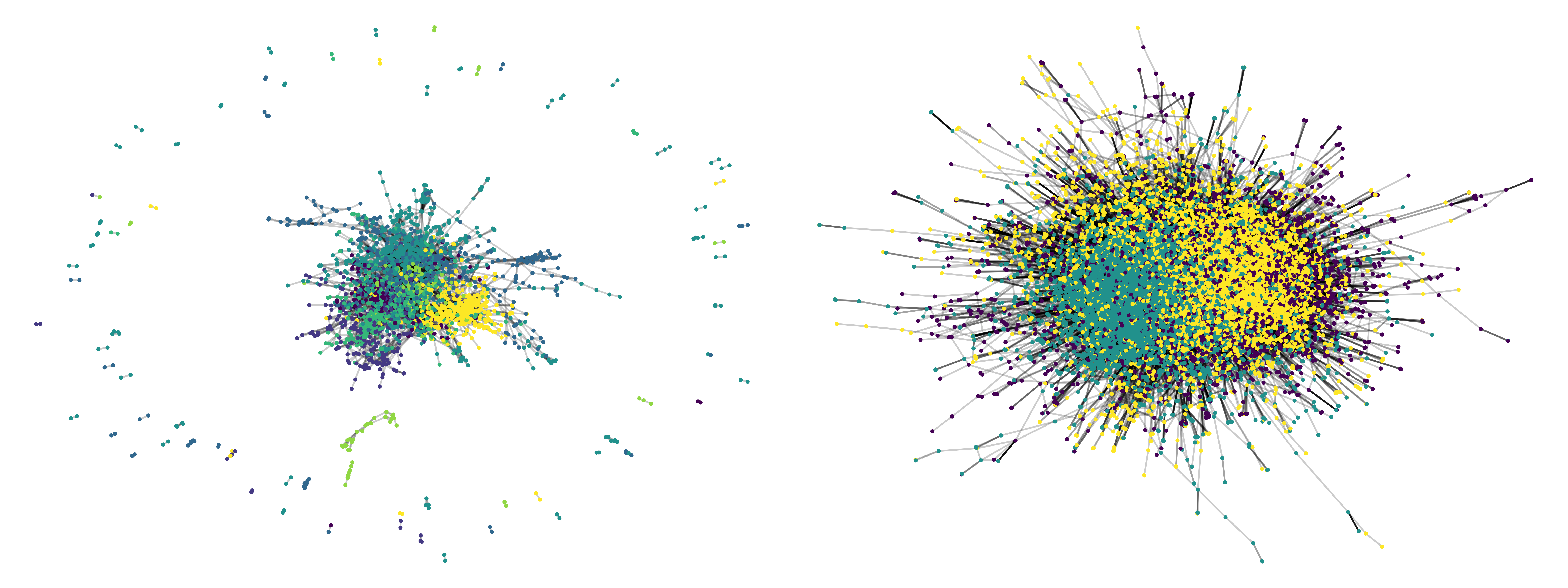}
  \caption{Visualization of Cora and Pubmed networks, colored based on node labels.}
  \label{fig:networks}
\end{figure}

The datasets used for the evaluation of the embedding algorithms consist of 11 real-world complex networks. The summary of the datasets is shown in Table~\ref{tab1}. This table shows that we use datasets that have different characteristics since they differ a lot in the number of nodes, edges, classes, and connected components. We also show visualizations of Cora and Pubmed datasets in Fig.~\ref{fig:networks}. In the figure, target classes are represented using different colors.

\begin{itemize}
\item \emph{Ions}~(\cite{skrlj2019deep,minfskrlj}) is a network of ion binding sites, linked by their structural similarity. The target class is the type of ion that binds to a given protein substructure (node).
\item \emph{Cora}~(\cite{Qing2003citeseer}) is a network of scientific publications and the citations between them. The labels represent the topic categories of the publication.
\item \emph{CiteSeer}~(\cite{Qing2003citeseer}) is a network of scientific publications and the citations between them. The labels represent the topic categories of the publication.
\item \emph{Bitcoin Alpha}~(\cite{skrlj2019deep}) is a network of Bitcoin transactions from the platform Bitcoin Alpha. The labels represent the level of trust in the transaction (integer range from -10 to 10).
\item \emph{Homo sapiens (PPI)} (as used in~(\cite{grover2016node2vec})) is a network of the proteome, i.e. a set of proteins which interact with each other. The labels represent the protein functions.
\item \emph{Wikipedia}~(\cite{mahoney2011large}) is a network of co-occurrences of words in the first million bytes of the Wikipedia dump. The labels represent the part-of-speech tags.
\item \emph{Bitcoin}~(\cite{skrlj2019deep}) is a network of Bitcoin transactions from the platform Bitcoin OTC. The labels represent the level of trust in the transaction (integer range from -10 to 10).
\item \emph{BlogCatalog}~(\cite{zafarani2009social}) is a network of social relationships on the Blogger website. The labels represent interests inferred from metadata provided by the authors. 
\item \emph{Coauthor-CS}~(\cite{oleks2018pitfalls}) is a computer science co-authorship network where nodes represent authors and edges represent that two authors co-authored a paper. The labels represent the authors most active fields of study.
\item \emph{Pubmed} (as used in~(\cite{wang2020unifying})) is a network of scientific publications and the citations between them. The labels represent the topic categories of the publication.
\item \emph{Coauthor-PHY}~(\cite{oleks2018pitfalls}) is a physics co-authorship network, where nodes represent authors and edges represent that two authors co-authored a paper. The labels represent the authors' most active fields of study.
\end{itemize}

\begin{table}
\centering
\caption{Basic statistics of the networks used for testing.}
\begin{tabular}{lrrrr}
\hline
Name & Nodes & Edges & \makecell{Connected \\ Components} & Classes\\
\hline
Ions&   1969&   16092&  326&    12\\
Cora&   2708&   5278&   78& 7\\
Citeseer&   3327&   4676&   438&    6\\
Bitcoin Alpha&    3783&   14124&  5&  20\\
Homo sapiens (PPI)&   3890&   38739&  35& 50\\
Wikipedia&    4777&   92517&  1&  40\\
Bitcoin&    5881&   21492&  4&  20\\
BlogCatalog&    10312&  333983& 1&  39\\
Coauthor-CS&    18333& 100227& 1& 15\\
Pubmed& 19717&  64041&  1&  3\\
Coauthor-PHY&    34493& 282455& 1& 5\\
\hline
\end{tabular}
\label{tab1}
\end{table}

\subsection{Experimental setting}
The conducted experiments focus on the multi-label node classification task. In the multi-label classification task, many different labels may be assigned to each instance. This can be formally looked at as a problem where we search for a function $\textrm{f}: N \rightarrow \{0,1\}^g$, where $g$ is the number of labels. Labels where the value is 1, are assigned to the instance, whereas the ones with 0 are not. An example of such a task is the assignment of genres to a movie.

When comparing the proposed method to the baselines, we evaluated the performance of a given embedding algorithm with the same methodology as in state-of-the-art papers such as node2vec~(\cite{grover2016node2vec}). The methodology is described below.

\begin{itemize}
\item We embedded a network's nodes to a low-dimensional representation.
\item We made ten copies of the embedding with corresponding labels and shuffled each.
\item We evaluated the performance on each copy using a training set of increasing size, i.e. from 10\% to 90\% classified by logistic regression. We classified each node into top $k_i$ classes based on the probability returned from the classifier, where $k_i$ represents the number of classes of a given node.
\item We calculated micro and macro F1 scores and averaged the results for each percentage.
\item We performed the described test for each embedding algorithm ten times.
\end{itemize}

The exception to this method of testing is the Label Propagation algorithm that does not use an embedding. To test it we ran the algorithm 100 times with the randomly selected training set of increasing size from 10\% to 90\%, similarly to how we tested the other embedding algorithms.

All experiments were conducted on a machine with 128 GB RAM, Intel(R) Xeon(R) Gold 6150 @ 2.7 GHz with a NVIDIA Tesla V100 SXM3 32 GB GPU. The approaches that consumed more than 128 GB of RAM were marked as unsuccessful and are shown as Out Of Memory (OOM) in the results. We added this constraint because we use medium-sized datasets for testing and the methods that need more memory would probably not scale well to larger networks.

As default parameters for SNoRe we use $\epsilon = 0.005$, maximum walk length $=5$, number of walks per node $= 1024$ and $2048$ pivot nodes ($d$). For SNoRe (SDF) we use the same parameters except that we use $d$ equivalent to a dense representation with 256 features ($\tau = |N|\cdot 256$). We have chosen 256 features because other embedding algorithms we tested use 32-bit floating-point numbers with 128 features whereas we use 16-bit floating-point values, making the size of the embedding the same.

\subsection{Baselines}
We compared the results of the proposed approach against the results of eight other baselines outlined below. Seven of these are embedding algorithms, the exception being Label Propagation that performs classification directly by propagating label information across the network structure.

\begin{itemize}
\item \emph{Random baseline} creates an embedding of size $|N|\times 64$ with random numbers drawn from $\textrm{Unif}(0,1)$.
\item \emph{Label Propagation (LP)}~(\cite{zhu2002learning}) propagates labels of annotated nodes through the network until convergence or the maximum number of iterations. We used alpha $= 0.9$ as parameter.
\item \emph{VGAE}~(\cite{kipf2016variational}) is a variational auto-encoder that uses latent variables to learn a model that can be interpreted. This auto-encoder is used mostly for link prediction. We used default parameters epochs $=200$, learning rate $=0.01$, $32$-dim hidden layer and $16$-dim latent variables in the experiments.
\item \emph{Personalized Page Rank with Shrinking (PPRS)}. This variant of Personalized PageRank was developed as part of HINMINE methodology~(\cite{kralj2017hinmine}). The algorithm, for each node, computes its representation by iteratively obtaining a discrete stationary distribution of walk visits. The shrinking offers additional speedups. We use probability threshold $= 0.0005$ and number of important $= 1000$ that are the default parameters for testing.
\item \emph{DeepWalk}~(\cite{Perozzi2014deepwalk}) equates random walks to sentences. These sentences are used to learn the network representation using a simple language model-like procedure. We use default parameters: representation size $=128$, walk length $=80$, and the number of walks $= 10$ in the experiments.
\item \emph{NetMF (SCD)}~(\cite{skrlj2019embeddingbased}) is the PyTorch~(\cite{paszke2017automatic}) re-implementation of the NetMF embedder~(\cite{qiu2018network}). NetMF tries to approximate the closed form of the DeepWalk's implicit latent walk matrix. The re-implementation is suitable for highly sparse matrices and is optimized for running on GPUs, offering substantial performance improvements. We use the default parameters: dimension $= 128$, window size $= 10$, rank $= 248$ and negative $= 1$ in the experiments.
\item \emph{LINE}~(\cite{tang2015line}) is one of the first network embedding algorithms. It uses an objective function that preserves first and second-order proximities. We use default parameters: embedding dimension $= 200$ and the number of negative samples $= 5$ in the experiments.
\item \emph{node2vec}~(\cite{grover2016node2vec}) learns a low dimensional representation of nodes that maximizes the likelihood of neighborhood preservation using random walks. We use default parameters: embedding dimension $= 128$, walk length $= 80$, number of walks $= 10$ and window size $= 10$ in the experiments.
\end{itemize}

\section{Results}
\label{sec:results}
We next present the results of the empirical evaluation. We begin with the classification results across the considered real-life datasets, followed by a series of ablation studies, where we explored SNoRe's behaviour in more detail, ranging from its explainability capabilities to behaviour w.r.t. different hyperparameter settings.

\subsection{Classification results}
\label{sec:classification}
\begin{figure}[t!]
  \centering
  \includegraphics[width = \linewidth]{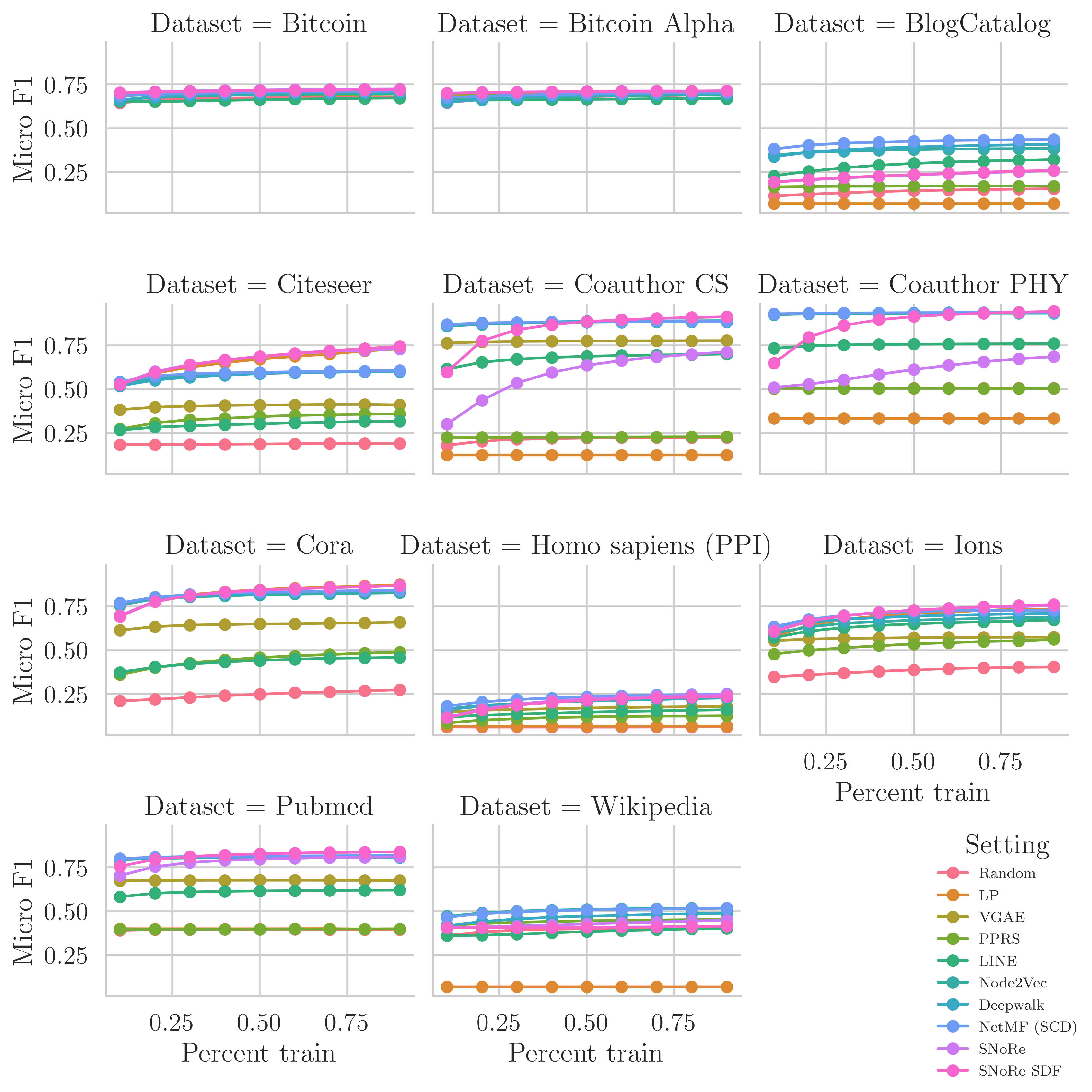}
  \caption{Micro F1 plots.}
  \label{fig:micro}
\end{figure}

\begin{figure}[t!]
  \centering
  \includegraphics[width = \linewidth]{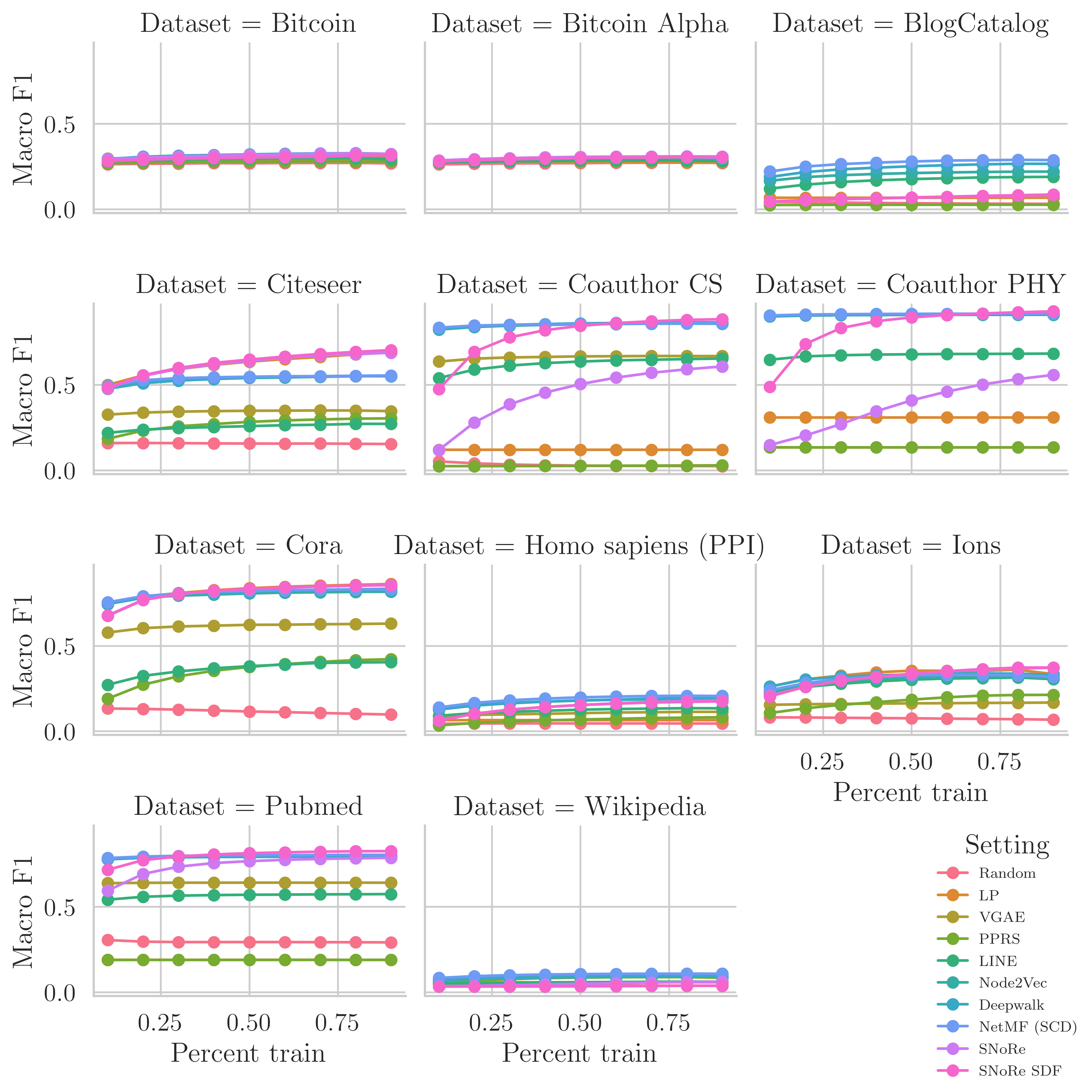}
  \caption{Macro F1 plots.}
  \label{fig:macro}
\end{figure}

Classification results are visualized in Fig.~\ref{fig:micro} and~\ref{fig:macro}, as well as presented in tabular format, where average performances across different training percentages alongside the corresponding standard deviations are reported (Tables~\ref{tab-micro} and~\ref{tab-macro}).

It can be observed that the proposed SNoRe algorithm performs competitively, or even outperforms the considered baselines. We can see that SNoRe and its extension SNoRe (SDF) work well on co-authorship networks, citation networks, Cora and the Ions dataset. Their results on both co-authorship network are interesting, since F1 scores are low at first, but then they rise fast and achieve the best results out of all baselines when we use enough training instances. Our algorithm performs poorly compared to other baseline methods on datasets such as Wikipedia and BlogCatalog, where nodes with similar class do not necessarily have similar neighborhoods (homophily), which is potentially the case with the co-authorship datasets. We can see that all embedding algorithms perform similarly to the random baseline on both Bitcoin datasets. This potentially shows that some datasets may not be suitable for direct learning.

\begin{table}[htb!]
    \centering
	\input{mezna.t1}
	\caption{Mean aggregated micro F1 scores.}
	\label{tab-micro}
\end{table}
\begin{table}[htb!]
    \centering
	\input{mezna.t2}
	\caption{Mean aggregated macro F1 scores.}
	\label{tab-macro}
\end{table}

Similar results can be observed in the averaged results (Tables~\ref{tab-micro} and~\ref{tab-macro}), indicating SNoRe and its extension offer the state-of-the-art performance, albeit offering fundamentally different representation learning capabilities (sparse and symbolic). The results of both SNoRe and SNoRe (SDF) in the averaged tables suffer because the classification accuracy drops when we use a small amount of data (less than 25\%) for training. The table also shows that SNoRe performs the best on four datasets (same amount as NetMF (SCD)) in the micro F1 metric while only achieving the best result on one dataset in the macro F1 metric.

\subsection{Statistical analysis}
\label{sec:statistical}
This section presents the statistical comparison of embedding algorithms by using average rank diagrams with critical distances~(\cite{demsar2006statistical}) and also Bayesian comparisons~(\cite{benavoli2017bayes}).

\begin{figure*}[t!]
  \centering
  \includegraphics[width = \linewidth]{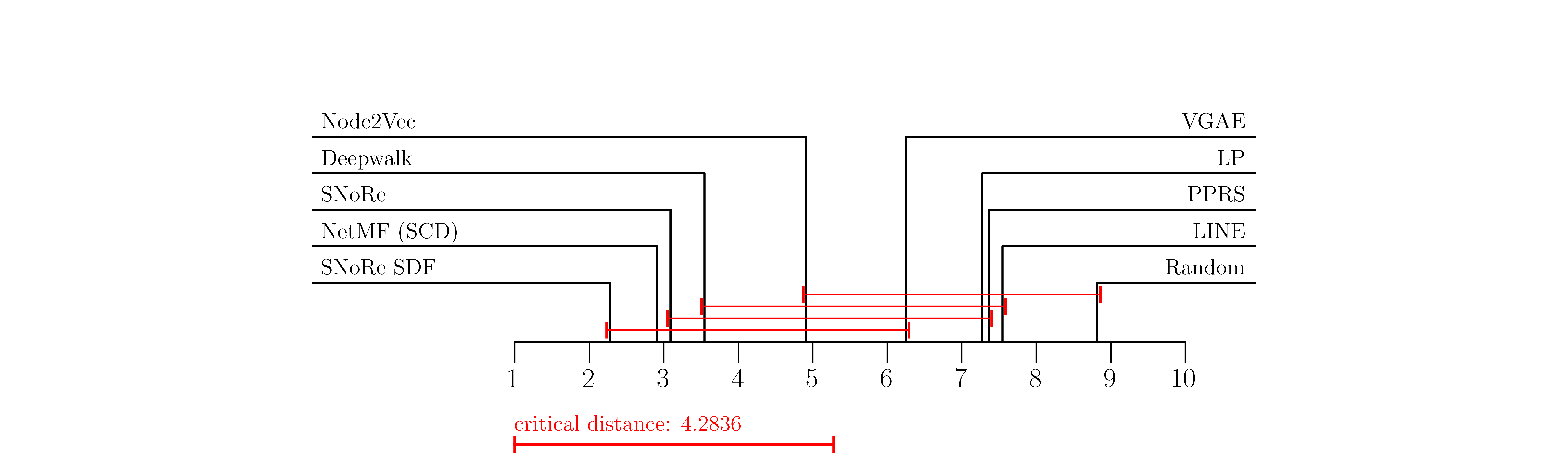}
  \caption{Micro F1 average rank diagram where best performing percentage is chosen.}
  \label{fig:cdmicromax}
\end{figure*}

\begin{figure*}[t!]
  \centering
  \includegraphics[width = \linewidth]{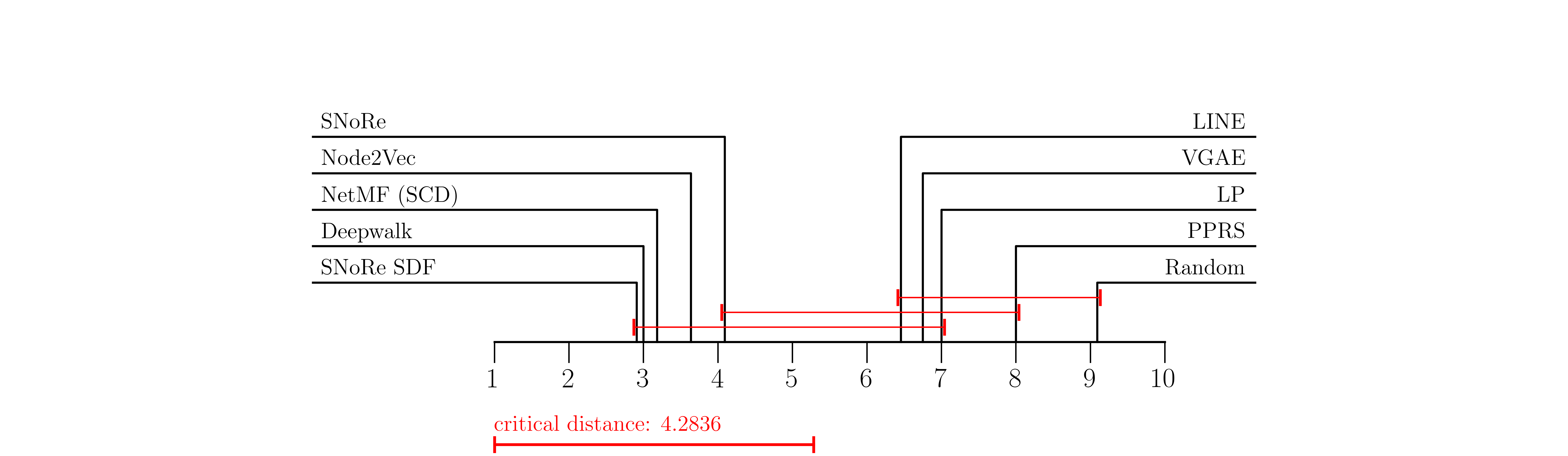}
  \caption{Macro F1 average rank diagram where best performing percentage is chosen.}
  \label{fig:cdmacromax}
\end{figure*}

Average rank diagrams are shown in Fig.~\ref{fig:cdmicromax} and~\ref{fig:cdmacromax}. These diagrams display the mean rank of algorithms over all datasets along the horizontal line. The ranks used in these diagrams are assigned to the algorithms based on their best performing percentage on a given dataset. We assigned ranks in this way because we usually only want to classify a few new instances using models trained on larger amounts of labeled data. More diagrams showing the performance where ranks are assigned based on mean results over all percentages on a dataset can be found in Appendix~\ref{sec:app-rank}. The critical distance in the figures group classifiers whose rank is not significantly different (Friedman test with Nemenyi \emph{post-hoc} correction).

We see that for both micro and macro F1 metric SNoRe (SDF) performs best out of all algorithms and that when a constant amount (2048) of pivot nodes are used, SNoRe performs observably worse being fifth overall in the macro F1 metric. We can see that NetMF (SCD), DeepWalk, node2vec, and both versions of SNoRe form a group of algorithms that are state-of-the-art and perform significantly better than the other embedding algorithms. Our claim that SNoRe performs similar to other state-of-the-art algorithms is further backed by the critical distance that groups both versions of SNoRe with node2vec, DeepWalk, and NetMF (SCD).

Bayesian variants of performance comparison classifiers were recently introduced as a way to combat the shortcomings of methods like null hypothesis significance testing (NHST)~(\cite{benavoli2017bayes}). We use the Bayesian variant of the hierarchical t-test to determine differences in performance of compared classifiers. This test distinguishes between three scenarios: two where one of the classifiers outperforms the other and the one in which the difference in classifier performance lies in the region of practical equivalence (rope). The size of rope is a free parameter set to $0.01$ in our experiments, which means that two performances are considered the same if they differ by less then $0.01$. An algorithm can be argued to perform significantly better, if $p(\textrm{algorithm})>0.95$.

\begin{figure}[t!]
  \centering
  \includegraphics[width = \linewidth]{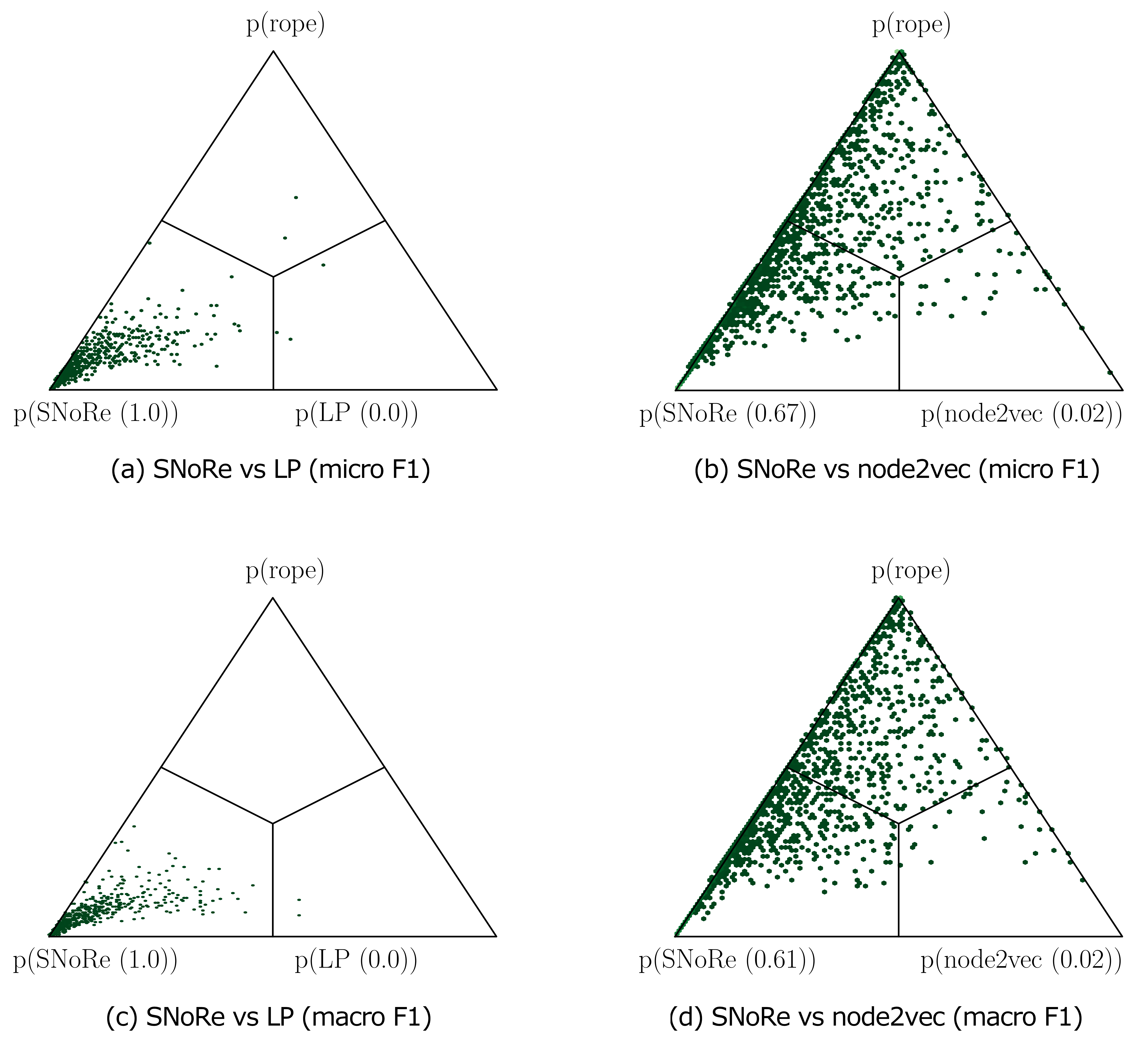}
  \caption{Pairwise Bayesian performance comparisons of selected classifiers. The probabilities following classifier names represent the probabilities that a given classifier outperforms the other.}
  \label{fig:bayes}
\end{figure}

As Bayesian multiple classifier correction cannot be intuitively visualized for more than two classifiers, we show the comparison between SNoRe (SDF) and node2vec as well as Label Propagation in  Fig.~\ref{fig:bayes}. The two comparisons are used to demonstrate the performance against a strong and a weak baseline. We chose node2vec as the strong baseline because it is a generalization of Deepwalk and thus covers both algorithms. The data used to make these comparisons was collected over all datasets using ten repetitions of ten-fold cross-validation.

The green dots in the triangles represent samples, obtained from the hierarchical model. As the sampling procedure is governed by the underlying data, green dots fall under one of the three categories; classifier one dominates (left), classifier two dominates (right), or the difference of the classifiers' performance lies in the region of practical equivalence (up). Upon model convergence, some areas of the triangle are more densely populated, showing higher probability that the classifier outperformed the other. We can see that in our experiment SNoRe (SDF) significantly outperformed the Label Propagation algorithm in both micro and macro F1 metric, having almost all green dots in the far left corner. More interesting are the comparisons against node2vec where SNoRe still outperforms node2vec whose probability of outperforming SNoRe is only 2\%. Here a lot of dots are in the region of practical equivalence showing that both algorithm perform similarly a lot of times.

\subsection{Ablation study - parameter space exploration}
Having shown that the default hyperparameter setting $\epsilon = 0.005$, maximum walk length of $5$, number of walks $= 1024$ and $2048$ pivot nodes performs competitively to state-of-the-art, we conducted additional experiments to better understand SNoRe's behaviour w.r.t. different parameter settings. In this section, we present only plots of micro F1 performance. Additional plots with macro F1 results can be found in  Appendix~\ref{sec:app-macro}.

As the default hyperparameter value of $\epsilon$, we chose 0.005 because we wanted hashes with at most 200 non-zero values.

\subsubsection{Representation dimension}
\label{sec:rep-dim}
\begin{figure}[t!]
  \centering
  \includegraphics[width = \linewidth] {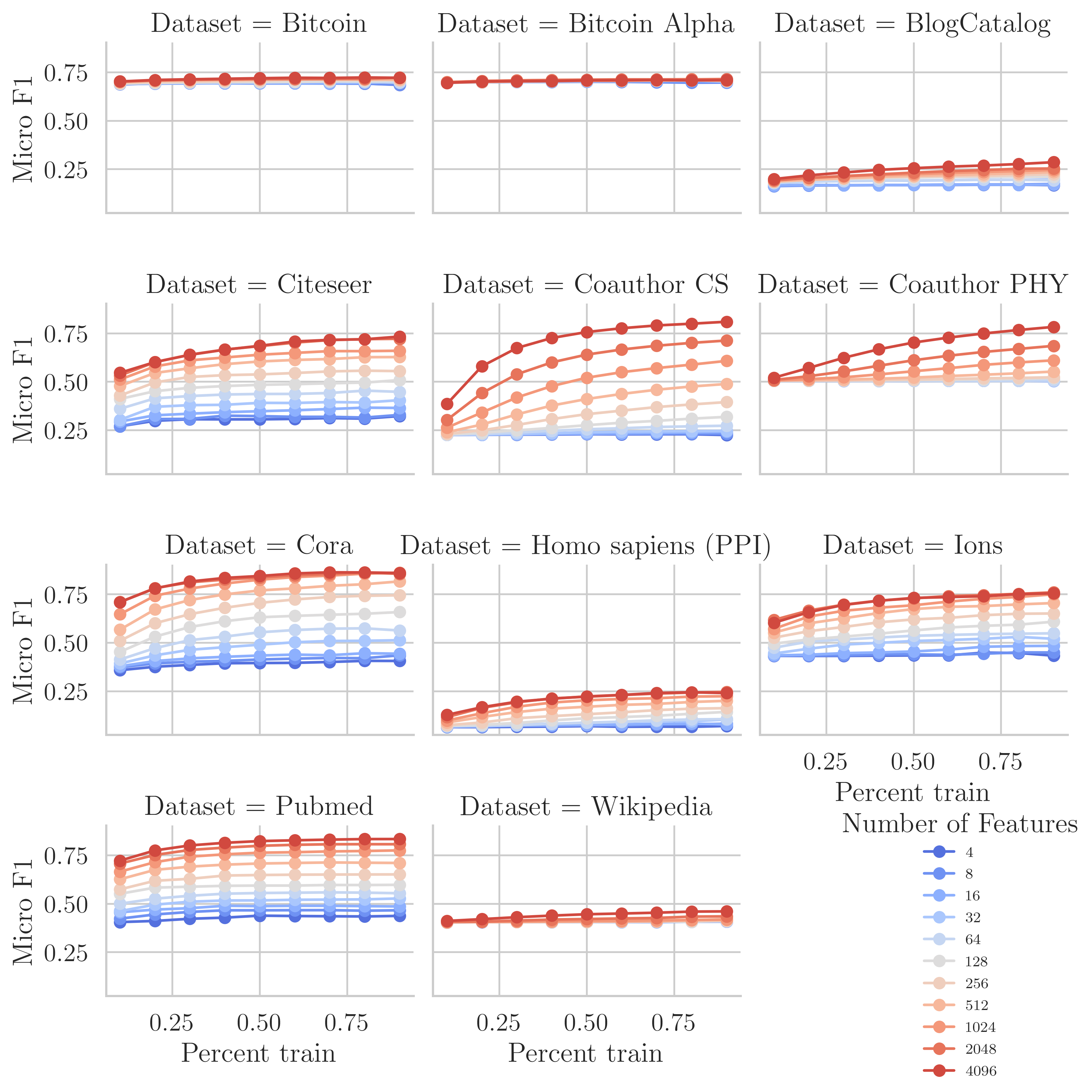}
  \caption{Micro F1 plots for different number of features.}
  \label{fig:microfeat}
\end{figure}

Fig.~\ref{fig:microfeat} shows the impact of different number of pivot nodes. From the figure, we can extract two types of datasets. Those where the score rises gradually and those where the score is similar no matter the number of features. Most tested datasets can be easily put in one of those groups, the exception being the BlogCatalog dataset and Homo Sapiens (PPI) dataset where the score rises slowly. Since the score noticeably rises on the Homo Sapiens dataset, we would put it in the first group, whereas we would put the BlogCatalog dataset into the second one. Coincidentally, BlogCatalog is the only dataset where SNoRe performs significantly worse than some other state-of-the-art methods. From Fig.~\ref{fig:micro} and~\ref{fig:macro} we can further observe that the results on datasets where the number of features does not affect the score are usually similar no matter which embedder we use and close in many cases to those of the random baseline. These might mean that these datasets are less susceptible to classification. 

\subsubsection{Maximum walk length}
\begin{figure}[t!]
  \centering
  \includegraphics[width = \linewidth] {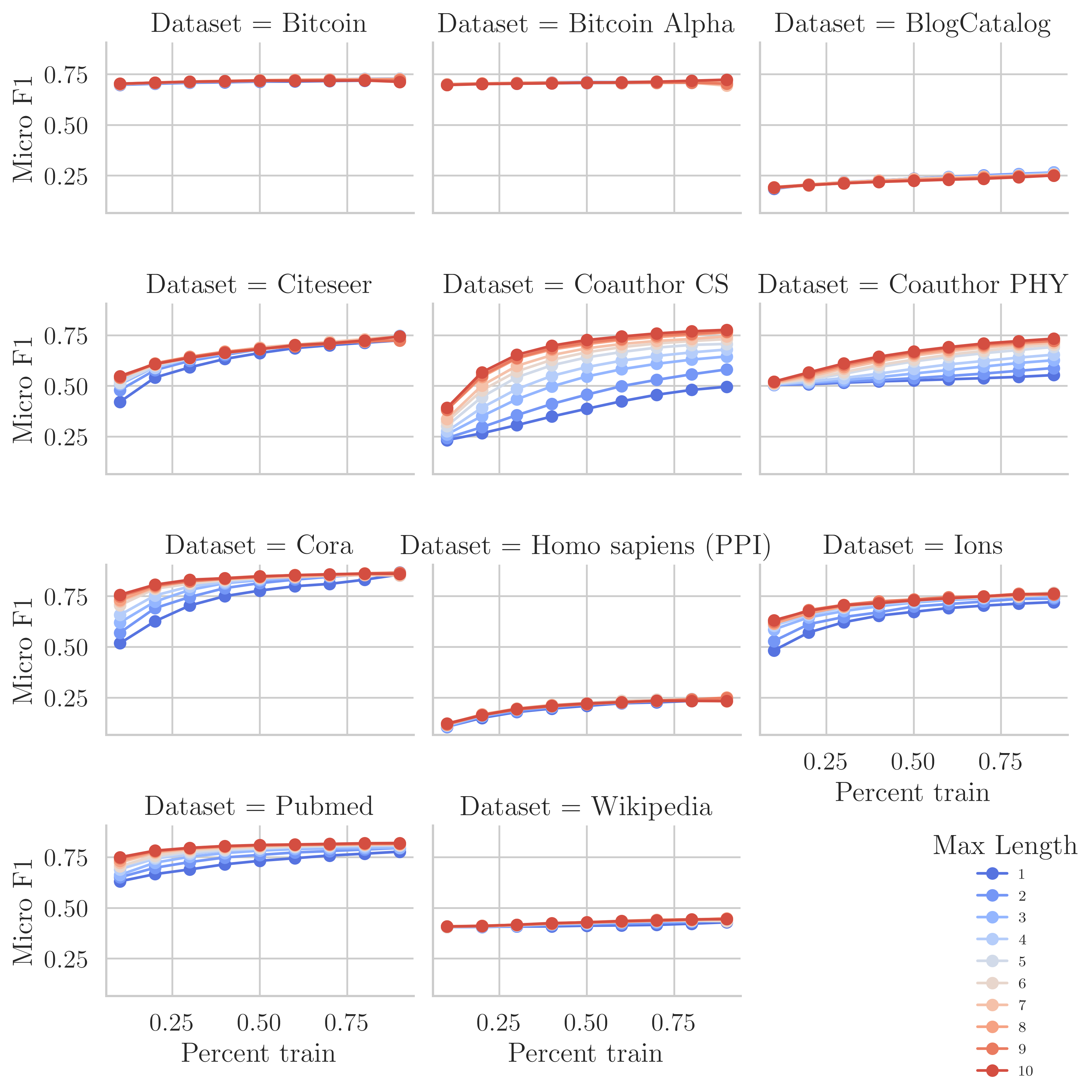}
  \caption{Micro F1 plots showing the effect of maximum walk length parameter.}
  \label{fig:microlength}
\end{figure}

We show the effect of the maximum walk length parameter in Fig.~\ref{fig:microlength}. From the figure, we can see that the score increases when we increase the value of this parameter. The increase of the score is especially apparent in the Coauthor PHY and Coauthor CS datasets. On the other hand, scores on Homo Sapiens (PPI), Wikipedia, BlogCatalog, and both Bitcoin datasets do not change much when the maximum walk length parameter is changed. This groups the datasets into almost the same groups as those in Section~\ref{sec:rep-dim}, with the exception being the Homo Sapiens dataset, where it was not clear to which group the dataset belongs.

We have chosen $5$ as the default value for the maximum walk length parameter. We have chosen this value because on most datasets, score does not significantly increase if we increase the parameter value, while the embedding time does. Execution time is explored further in Appendix~\ref{sec:app-time}.

\subsubsection{Number of random walks}

\begin{figure}[t!]
  \centering
  \includegraphics[width = \linewidth] {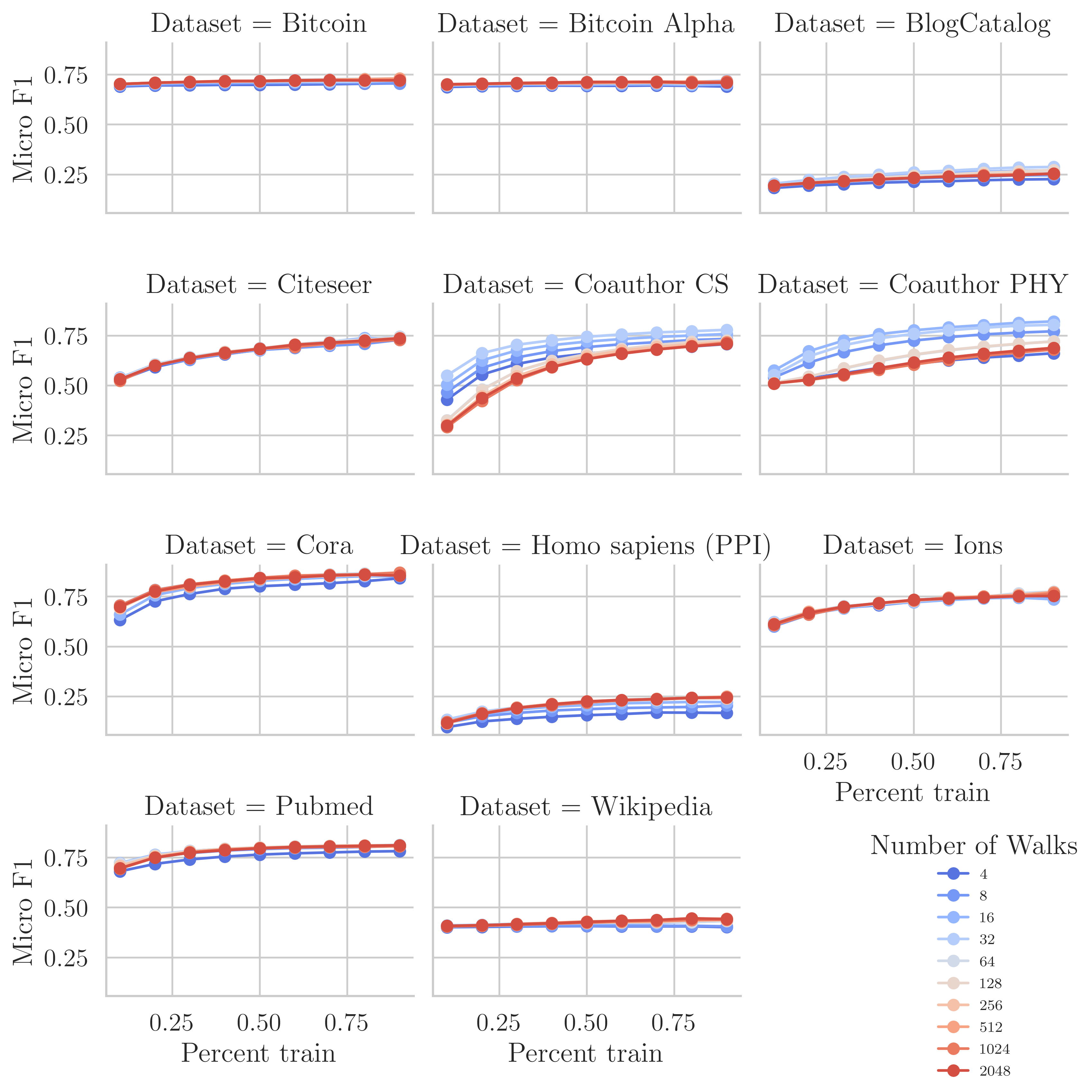}
  \caption{Micro F1 plots showing the effect of different number of random walks.}
  \label{fig:microwalks}
\end{figure}

We show the effect of the different number of random walks per node in Fig.~\ref{fig:microwalks}. We can see that the classification score on most datasets does not change much when different values of this parameter are chosen. The change is most observable on Coauthor CS and Coauthor PHY datasets, where the parameter values of 32-64 work best.

We have chosen $1024$ as the default value for the number of random walks per node parameter. We selected this value because it entails the maximum length needed across all datasets, and the execution is still fast. We also believe it is beneficial to create more walks since this gives us a greater chance to sample out relevant sub-graph structures.

\subsection{Ablation study - effect of different distances (metrics)}
\label{sec:distances}
In our approach, we selected cosine similarity to calculate the distance between two vectors because it offers good results and works well with sparse representation in both calculation and final embedding matrix. Since the choice of this distance metric is arbitrary we next show how the choice of distance metric affects the results. Table~\ref{tab:metrics} shows different distance metrics we compared using same, default parameters. In the formula for standardized Euclidean $v_i$ represents the variance of the $i$-th feature in the hash vector. These metrics where chosen because they represent different groups of distance metrics. Euclidean distance is a special case of Minkowski distance where $p=2$. It measures the distance between two points in the Euclidean space. By taking the variance of each dimension into account during the calculation of Euclidean distance, we get the Standardized Euclidean distance that is usually more robust when dimensions are scaled differently. Canberra distance is mostly used in intrusion detection and computer security and is a metric suitable for when the data is scattered around the origin. Jaccard similarity works on binary data and calculates similarity based on whether a feature is present or not. This metric can also be generalized for use with numeric values. The last metric we considered is the Hub Promoted Index (HPI) that was designed originally for quantifying the topological overlap between pairs of nodes in a network~(\cite{zhou2009hpi}). We generalized the metric, as shown in Table~\ref{tab:metrics}, to be used alongside the proposed hashing scheme.

\renewcommand{\arraystretch}{1.5}
\begin{table}
\centering
\caption{Used distance metrics and their formulas.}
\label{tab:metrics}
\setlength{\tabcolsep}{3pt}
\resizebox{.388\textwidth}{!}{
\begin{tabular}{cc}
\hline
Metric & Formula \\
\hline
Cosine&  \scalebox{0.99}{ $\frac{\sum_{i=1}^{|N|}\boldsymbol{a}_i\cdot \boldsymbol{b}_i} {\sqrt{ \sum_{i=1}^{|N|} \boldsymbol{a}_{i}^2} \sqrt{\sum_{i=1}^{|N|} \boldsymbol{b}_{i}^2}}$}\\
Euclidean&  \scalebox{0.99}{ $\sqrt{\sum_{i=1}^{|N|}(\boldsymbol{a}_i-\boldsymbol{b}_i)^2}$ }\\
\makecell{Standardized \\ Euclidean}& \scalebox{0.99}{$ \sqrt{\sum_{i=1}^{|N|} \frac{(\boldsymbol{a}_i - \boldsymbol{b}_i)^2}{\boldsymbol{v}_i^2}}$}\\
Canberra&   \scalebox{0.99}{ $\sum_{i=1}^{|N|} \frac{|\boldsymbol{a}_i| - |\boldsymbol{b}_i|}{|\boldsymbol{a}_i| + |\boldsymbol{b}_i|}$}\\
Jaccard&    \scalebox{0.99}{$ \frac{\sum_{i=1}^{|N|} \boldsymbol{a}_i \neq 0 \: \text{and} \: \boldsymbol{b}_i \neq 0}{\sum_{i=1}^{|N|} \boldsymbol{a}_i \neq 0 \: \text{or} \: \boldsymbol{b}_i \neq 0}$}\\
\makecell{Hub Promoted \\ Index (HPI)}&    \scalebox{0.99}{$ \frac{\sum_{i=1}^{|N|} \boldsymbol{a}_i \neq 0 \: \text{and} \: \boldsymbol{b}_i \neq 0}{min(\sum_{i=1}^{|N|} \boldsymbol{a}_i \neq 0, \sum_{i=1}^{|N|} \boldsymbol{b}_i \neq 0)}$}\\
\hline
\end{tabular}}
\end{table}
\renewcommand{\arraystretch}{1}

\begin{figure}[t!]
  \centering
  \includegraphics[width = \linewidth]{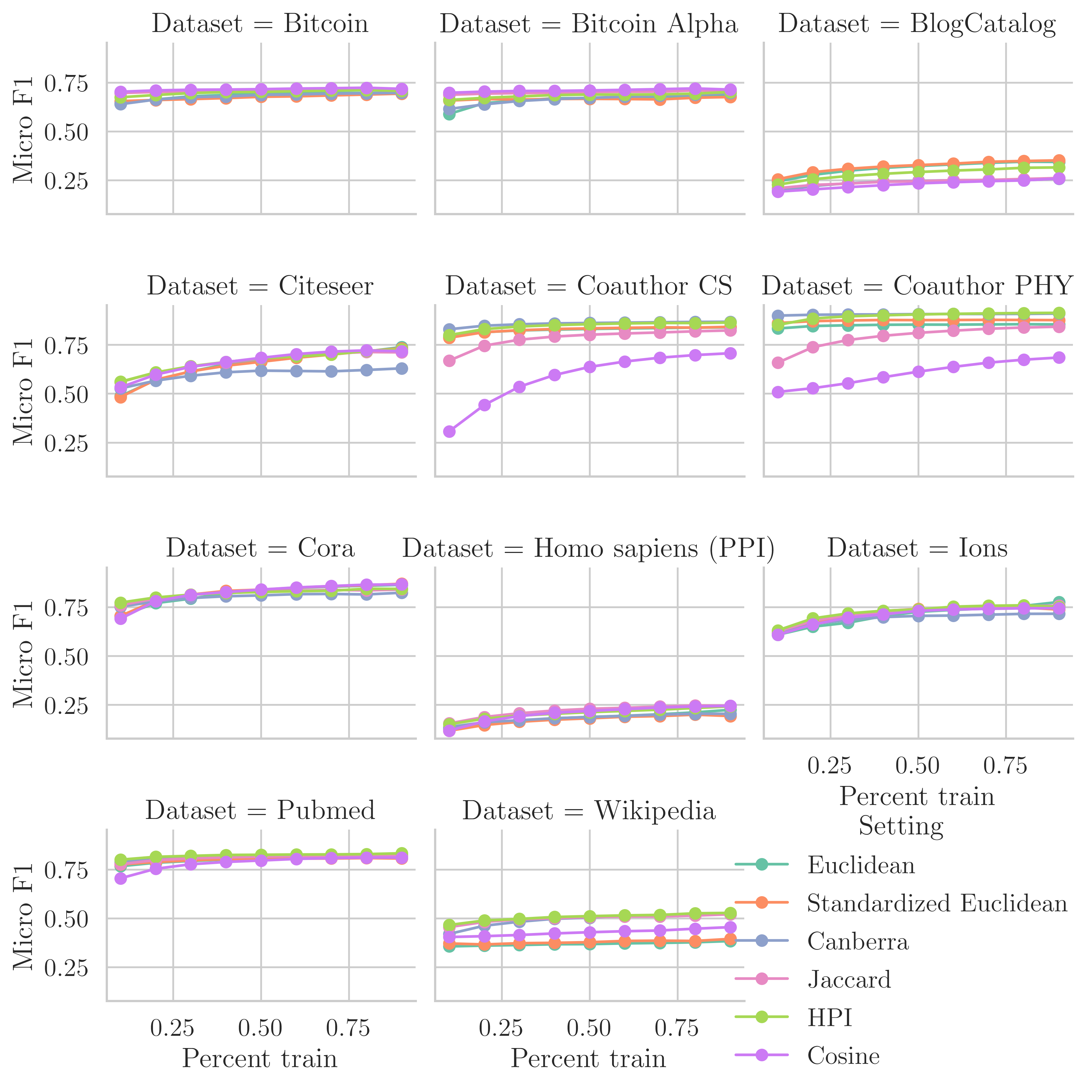}
  \caption{Micro F1 results for different distance metrics.}
  \label{fig:micrometric}
\end{figure}

\begin{figure}[t!]
  \centering
  \includegraphics[width = \linewidth]{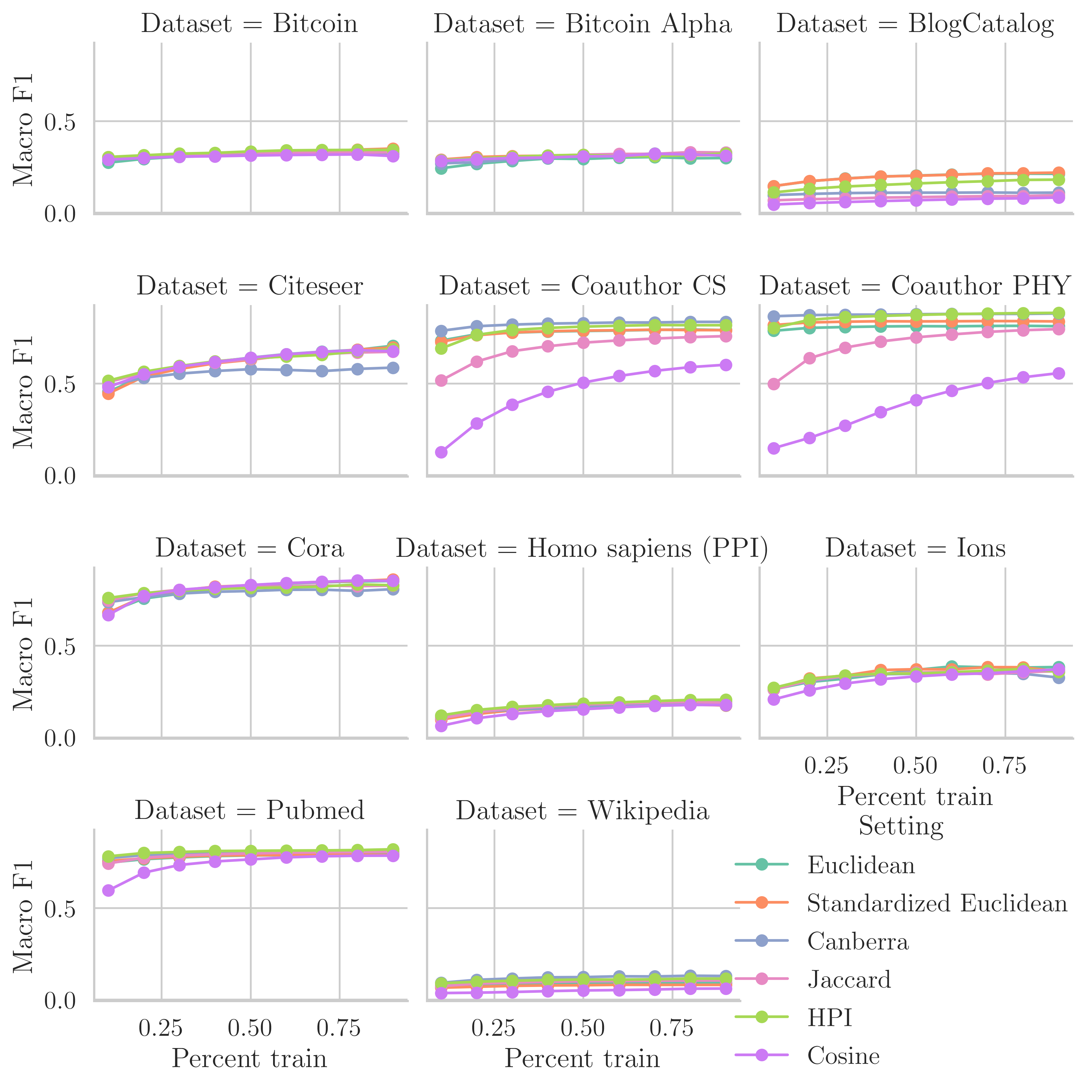}
  \caption{Macro F1 results for different distance metrics.}
  \label{fig:macrometric}
\end{figure}

The results between different distance metrics are shown in Fig.~\ref{fig:micrometric} and~\ref{fig:macrometric}. We can see that most metrics perform similarly on Bitcoin datasets, Citeseer, Cora, Homo sapiens, Ions and Pubmed. On BlogCatalog both Euclidean metrics and the HPI metric performed better than the other three. On both co-authorship datasets, cosine similarity performed worse than other metrics but the byte size of the embedding is significantly smaller since the embedding matrix is very sparse. Using SNoRe (SDF) where the size of representation is less than $\tau = |N|\cdot 128$ we get results that are better than those of other metrics. Using different distance metrics also helps on the Wikipedia dataset where the score is a lot higher for the Jaccard, Canberra and HPI metrics. As it should be expected both the Euclidean and Standardized Euclidean distance perform very similarly since SNoRe's hash function already normalizes the obtained values.

It should also be noted that cosine similarity, HPI and Jaccard similarities give us sparse embeddings, which perform significantly better when compared to the embeddings calculated using other metrics of the same size in bytes.

\subsection{Ablation study - evaluation time}
\label{sec:eval-times}
\begin{figure}[t!]
  \centering
  \includegraphics[width = \linewidth]{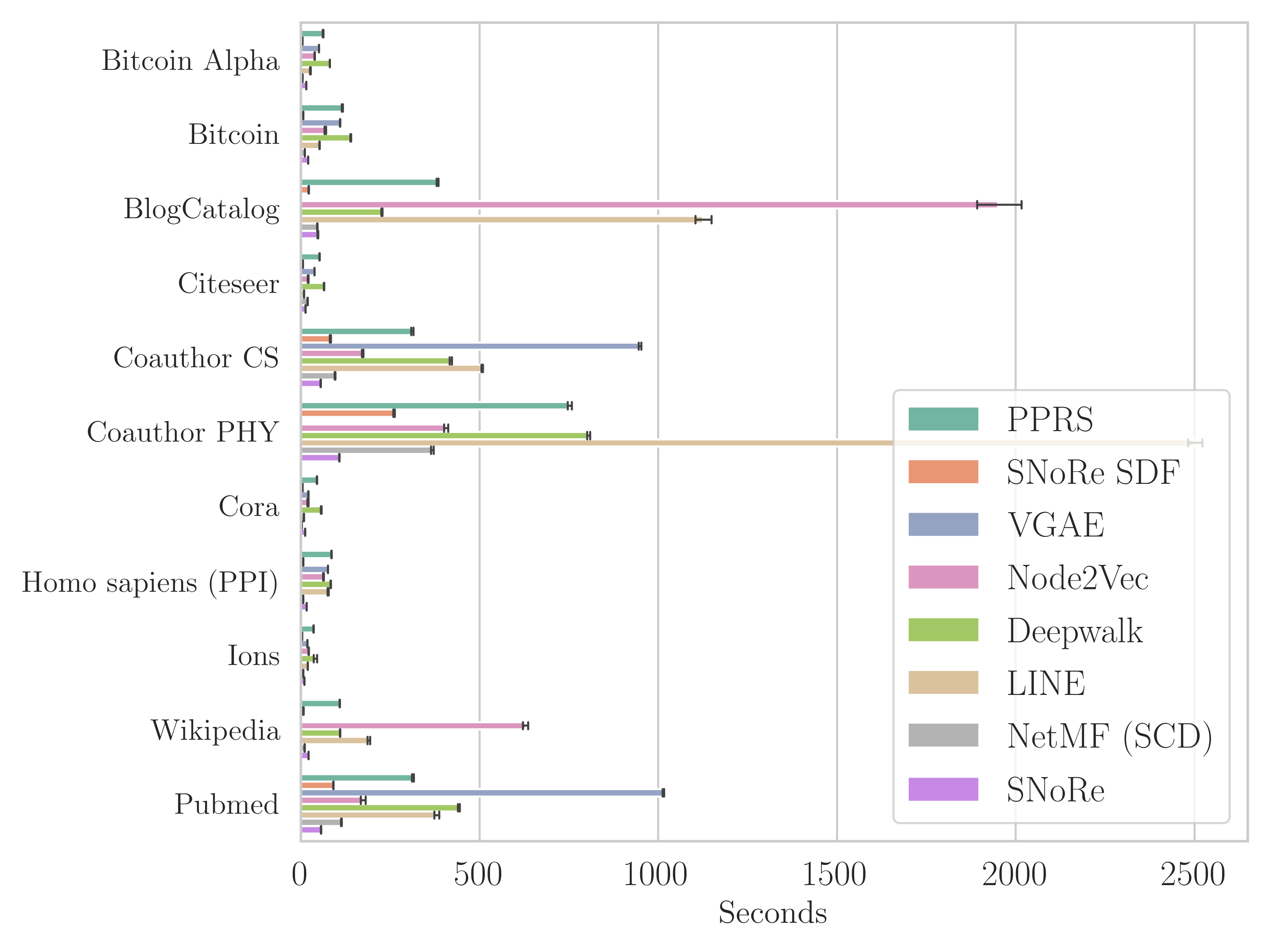}
  \caption{Visualization of execution time across the considered datasets.}
  \label{fig:time}
\end{figure}

In Section~\ref{sec:properties} we give the theoretical boundaries for time complexity. Here we give further empirical results and compare them to other baselines. The results between different baselines are shown in Fig.~\ref{fig:time}. We can see that both SNoRe and SNoRe (SDF) need a similar amount of time to finish and are usually the fastest, just before NetMF (SCD). We can see that SNoRe (SDF) is the fastest on small datasets but needs a little more time then SNoRe on larger datasets where more features need to be chosen.

A more extensive study of evaluation time for different number of pivot nodes, maximum walk length, number of random walks, and different distance metrics can be found in Appendix~\ref{sec:app-time}.

\subsection{Ablation study - explainability}
\label{sec:ablation-explainability}
In the final set of experiments, we demonstrate how SNoRe can be coupled with the existing model explanation approaches such as SHapley Additive exPlanations (SHAP)~(\cite{lundberg2017shap,vstrumbelj2014explaining}). SHAP is a game-theoretic approach used to explain any type of classification or regression model. The algorithm perturbs subsets of input features to take into account the interactions and redundancies between them. The explanation model can then be visualized, showing how the feature values of an instance impacted its classification.

\begin{figure*}[t!]
  \centering
  \includegraphics[width = 0.8\linewidth]{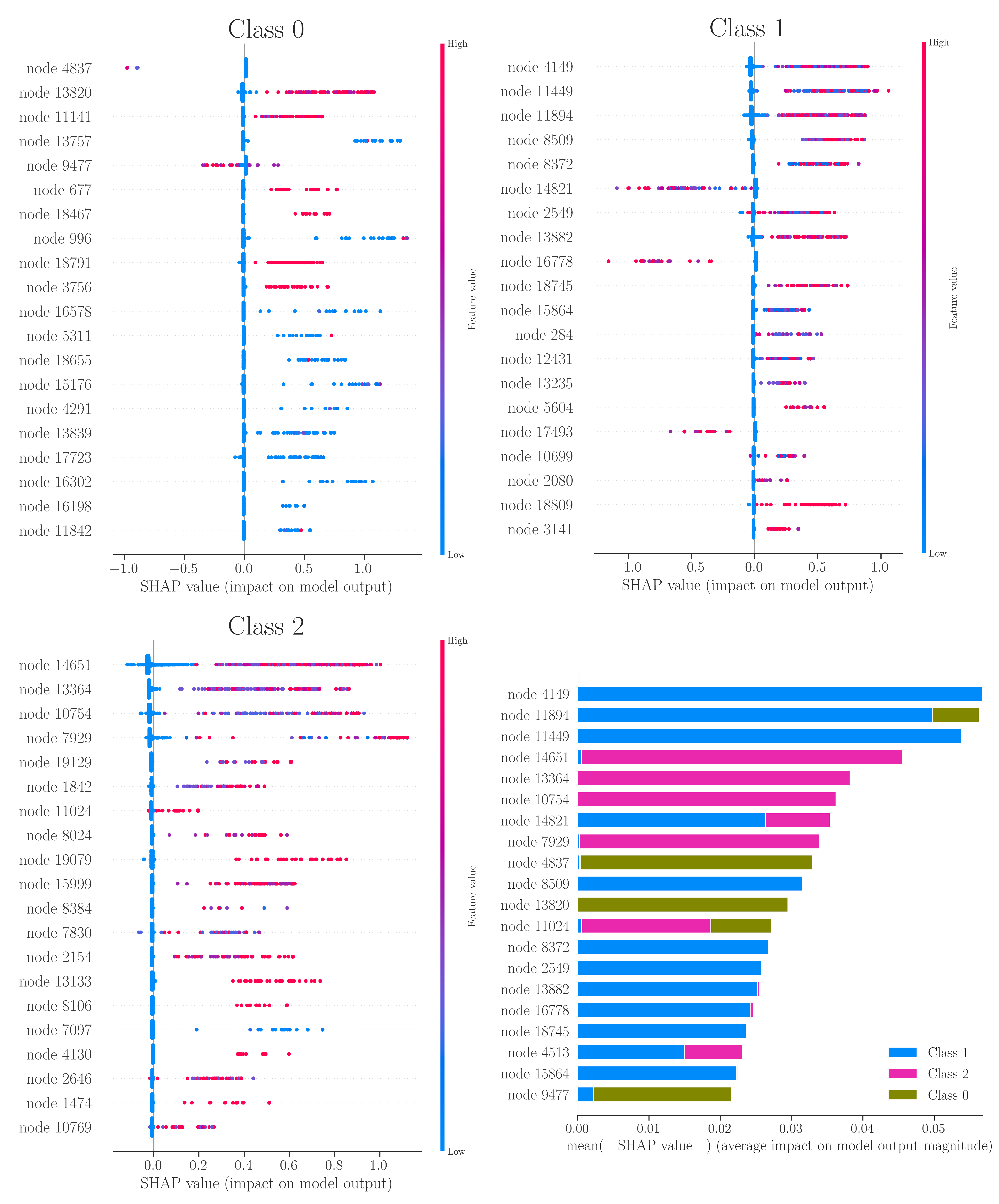}
  \caption{SHAP summary on Pubmed dataset.}
  \label{fig:shap}
\end{figure*}

We use the following methodology to explain how different feature values representing nodes impact how the classifier assigns a label to a node. This process will be showcased on the Pubmed dataset. First, we create the embedding and save indexes used as features. We then train the XGBoost model and input it to the SHAP tree explainer. We can then explain how different feature values impact an instance or create a summary of impact for all instances. With a summary, we can for example take the most impactful nodes, look at which articles they represent and look how they influence the assignment of classes. We created such a summary using SHAP library~(\cite{lundberg2020local2global}) and visualized the results in Fig.~\ref{fig:shap}. In the figure, the features are already renamed to indexes of the node (feature index $i$ is renamed to node $\textrm{feature-map}(i)$). Red and blue dots represent feature value, red being 1 and blue 0. We can see that usually, only high (non-zero) values impact how the model classifies a given instance since only those give information about nodes neighborhood. This can be seen in the figure, especially for the first three features of class 0. From the fourth feature in the summary table for class 0 (node 13757), we can see, that sometimes even low feature values (merely their presence) can have a big impact on the classification. The plot in the bottom right of Fig.~\ref{fig:shap} shows how much impact a feature has on average. We can see that node $4149$ has the biggest impact on classification of nodes and that usually when its value is high the node is classified to class 1.

Similarly, we can show which nodes impacted the classification of a single instance to explain why the node was classified as it was. This is further elaborated in Appendix~\ref{sec:app-shap}.

\subsection{Ablation study - latent clustering with UMAP}
\label{sec:umap}
\begin{figure}[t!]
  \centering
  \includegraphics[width = \linewidth]{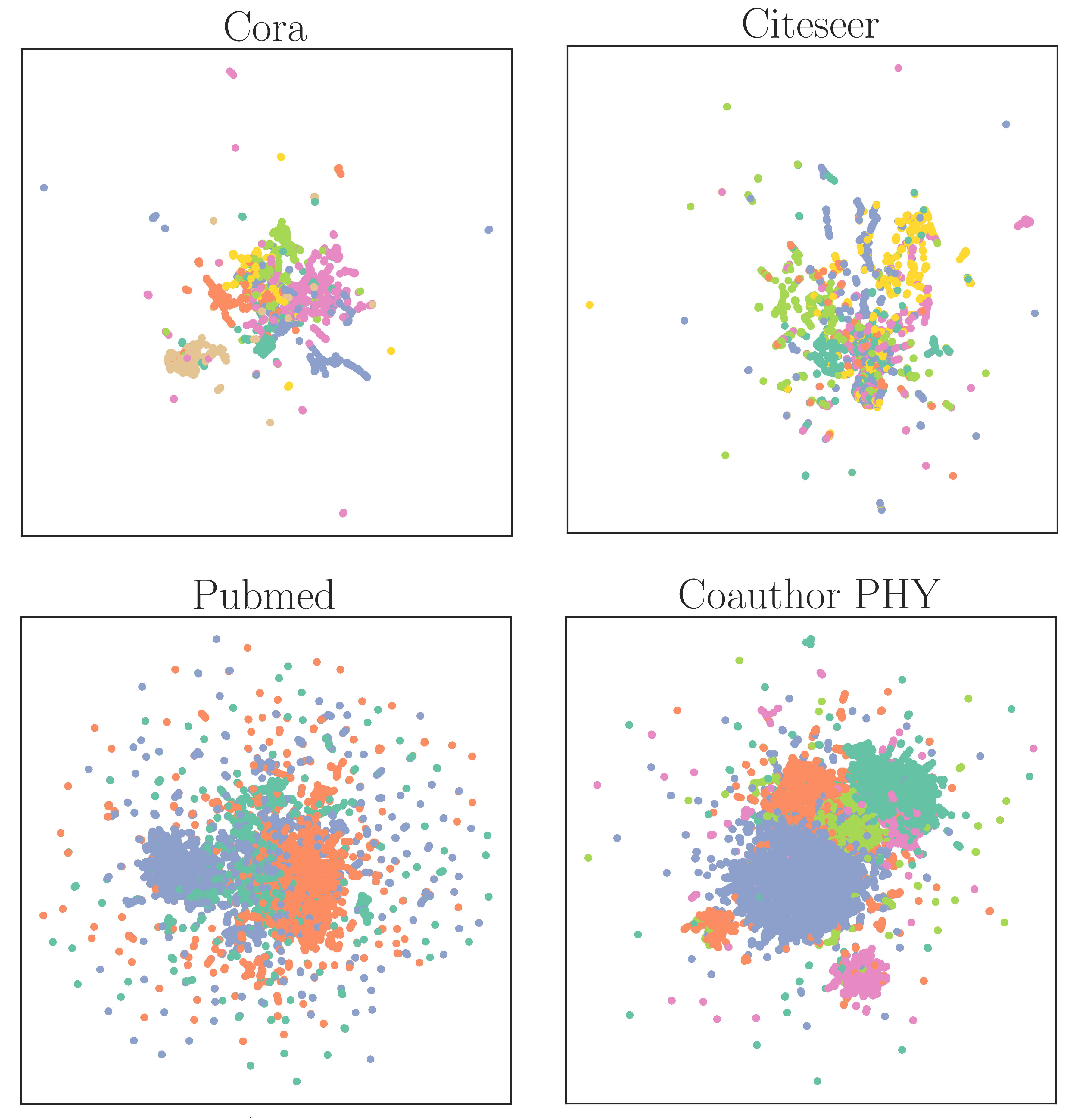}
  \caption{UMAP-based clustering of SNoRe (SDF) embedding on Cora, Citeseer, Pubmed and Coauthor PHY datasets.}
  \label{fig:umap}
\end{figure}

We also look at how nodes cluster together using UMAP algorithm~(\cite{lel2018umap}) to transform embedding vectors into 2D space. We saved the embedding of SNoRe (SDF) and used the default parameters for the unsupervised UMAP algorithm to generate node positions as shown in Fig.~\ref{fig:umap}. The class to which the node belongs to is shown as colour in the plot and added only for visualization. In general, we see that the nodes that belong to the same class are embedded near each other as best seen on the Coauthor PHY dataset. On the Pubmed dataset, we can observe that the classes coloured red and blue cluster well together and that the green one is scattered all over the plot, not clustering well. Nodes in the Cora embedding cluster well, but the classes are close together and sometimes overlap. The worst example we show is on dataset Citeseer where nodes do not cluster well and where classes overlap a lot, but some clusters can still be seen.

\section{Discussion}
\label{sec:discussion}
In this section we summarize the main results and their implications, and discuss the limitations of the proposed SNoRe approach.

As empirically shown in Section~\ref{sec:results}, SNoRe and SNoRe (SDF) outperform state-of-the-art methods on most datasets and perform comparably or slightly worse on others (e.g., Homo sapiens, Wikipedia). Coupled with the ability to use different distance metrics, \emph{speed} and \emph{explainability} of the embedding, this algorithm provides a very good alternative to the state-of-the-art algorithms. We further back this claim in Section~\ref{sec:statistical}, where we show that SNoRe outperforms the strong baseline node2vec according to the pairwise Bayesian performance comparisons.
In both execution time and classification results, we show that the proposed algorithm is scalable since it achieves best results on both the smallest and the largest dataset while using the same amount of space or less than the baselines we compared it to. This is further shown in Appendix~\ref{sec:app-time}, where the effects of parameters on execution time are shown. We also show the importance of efficient implementation of sparse algorithms and why such implementations are crucial for the future.

By observing the classification results of embeddings that have a different number of pivot nodes, maximum walk length and number of walks per node we have observed another interesting phenomenon. On datasets where all baselines achieved results that were similar to the random baseline, the parameters did not matter and for example an embedding with 4 features achieved similar results to the one with 4096. This gives us the ability to judge how susceptible a dataset is for classification and to judge if SNoRe is the suitable embedding algorithm for the task. 

While observing classification results between different parameters can give us an idea how susceptible a dataset is for classification, observing the number of features returned by SNoRe (SDF) can give us some insight into the structure of the network. This is most notable on the Wikipedia dataset where SNoRe (SDF) gives us a dense embedding since all nodes have at least one node in common. On the other hand, when using the same amount of space as a dense embedding on the Coauthor PHY dataset, our algorithm generates an embedding with all nodes used as features. This shows that the Coauthor PHY network is a lot more decentralized than the Wikipedia one.

SNoRe uses nodes as features, making it possible to explain the reasoning behind why an instance was classified in a certain way. This can be done with the use of tools such as SHAP and allows us to use this embedding algorithm in situations where explainability is crucial such as medicine.

In Section~\ref{sec:umap} we show that our algorithm creates an embedding that embeds nodes belonging to the same class, close together. We do this by using the UMAP algorithm to transform each instance into 2D space and by coloring the node w.r.t. the class they belong to. In the corresponding figure, we can easily see how nodes with the same class cluster on datasets Pubmed and Coauthor PHY and although a little less prevalent also on the other ones.    
Some of the limitations of our algorithm can be seen on datasets like Wikipedia and BlogCatalog, where the neighborhood of the node is not necessarily important and distinctive enough. Since the algorithm is modular this can probably be avoided sometimes by changing the hashing function is such a way that it better encodes the relevant network structure.

Although PageRank works very well on most networks, giving us features that give us good results, we cannot guarantee good features that span trough all the network will be chosen. This can drastically decreases the performance on some part of the network since some nodes may not have neighborhoods that overlap with the neighborhoods of the features.  

The last problem to highlight is the number of features (pivot nodes) in the final embedding. A small number of features is usually not descriptive enough and therefore the embedding performs badly. On the other hand, having a large number of features may give good results but need longer to train the classifier. Related to this, many classifiers are not optimized for sparse matrices.

\section{Conclusion and Further work}
\label{sec:conclusions}
We introduced a scalable unsupervised algorithm for learning symbolic node representations of networks. The algorithm is fast, achieves results that are comparable or better than those of state-of-art algorithms and can be interpreted when coupled with methods like SHAP.

This work offers extensive exploration, as well as a proof that symbolic representations, if learned based on a considered graph's global topology, offer a competitive paradigm to currently adopted black-box representation learning. The proposed SNoRe is freely available and highly optimized, and as such ready to be tested in many scenarios where black-box approaches are currently adopted -- for example in bioinformatics where the canonical task of protein function prediction is commonly addressed.

Further, the current version of SNoRe, offers symbolic representations of individual nodes which are in principle understandable and inspectable. However, we believe that, as domain knowledge in the form of e.g., ontologies is many times present, the obtained representations could be further \emph{generalized} in order to obtain even more informative descriptions of why a particular node has a given property of study. Hence, coupling SNoRe with existing work on relational reasoning will be explored as part of the future work.

The key focus of this paper revolved around the task of node classification. Analogous to how representations of links can be learned from e.g., DeepWalk embeddings, similar idea could be explored in the context of symbolic representations, offering explainable link prediction. Here, each link would be explained based on the presence of a particular collection of nodes in its neighborhood, offering a novel research venue applicable especially in high-risk scenarios such as the biomedical domain (for example, this ideas could be used to explore whether there is really an interaction between e.g., a pair of micro RNA molecules, and why this is the case?).

In further work, we plan to further explore how to incorporate different high-level network structures and the effect of different hashing functions. We also want to explore how different feature selection algorithms affect the performance and if the difference is significant when supervised algorithms are used. Another venue worth exploring is the use of different walk length distributions, which is not explored in this paper. In fields such as medicine, explainability of machine learning might not be enough for practical use. Because of this we want to further explore causal implications, as defined in~(\cite{Holzinger2019Causability}). We also want to research which metrics work best on different datasets (also known as metric learning). Lastly, SNoRe's behavior in the inductive and dynamic setting could be explored to further show the algorithms' usefulness.

\section*{Availability}
\label{sec:availability}
The proposed methodology is available as a Python library at: \url{https://github.com/smeznar/SNoRe}.

\bibliography{ms.bib}{}
\bibliographystyle{plain}

\appendix

\section{Comparison between digitized and non-digitized embedding}
\label{sec:app-dig}
\begin{figure}[t!]
  \centering
  \includegraphics[width = \linewidth]{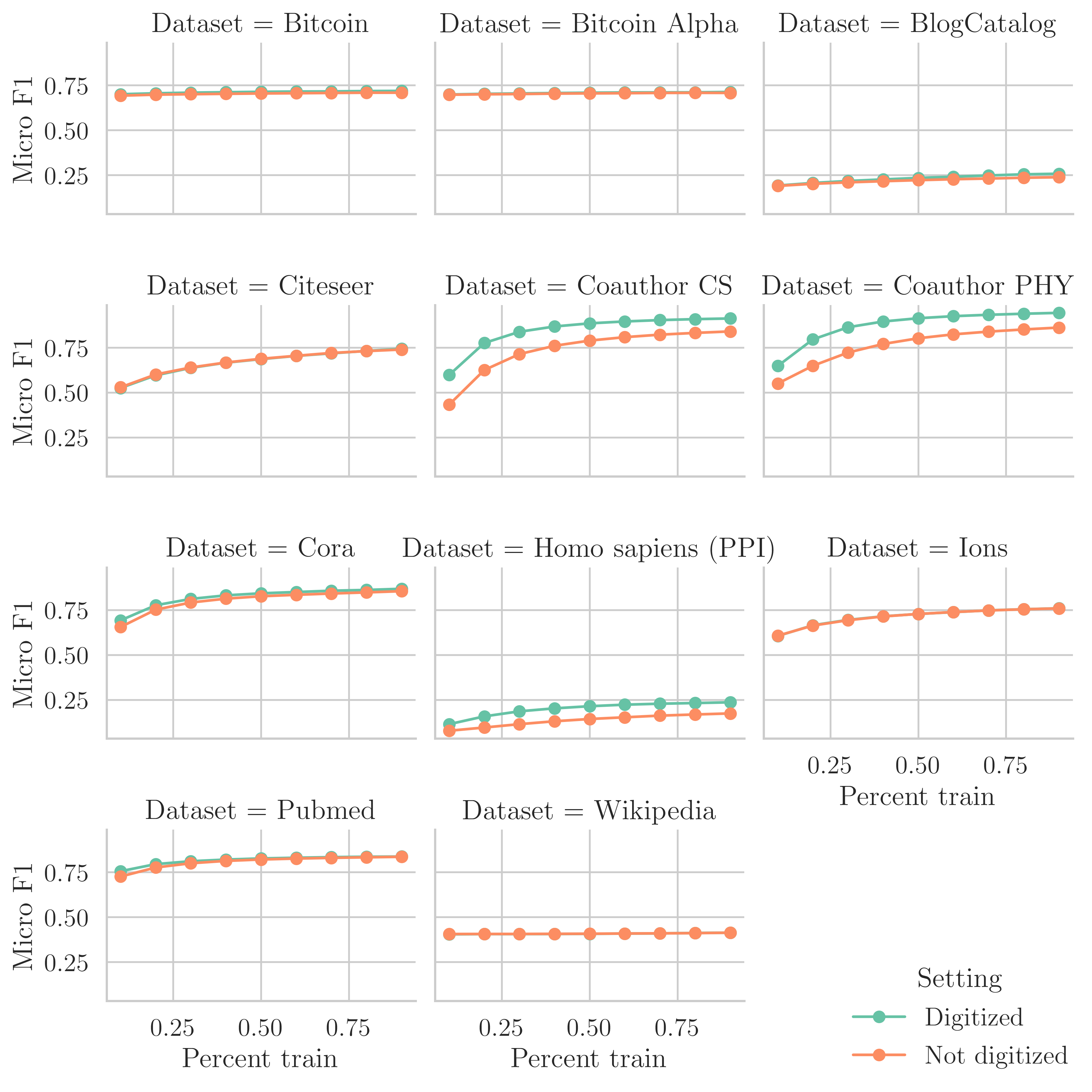}
  \caption{Micro F1 plots comparing digitized and non-digitized embedding.}
  \label{fig:microdigitized}
\end{figure}

\begin{figure}[t!]
  \centering
  \includegraphics[width = \linewidth]{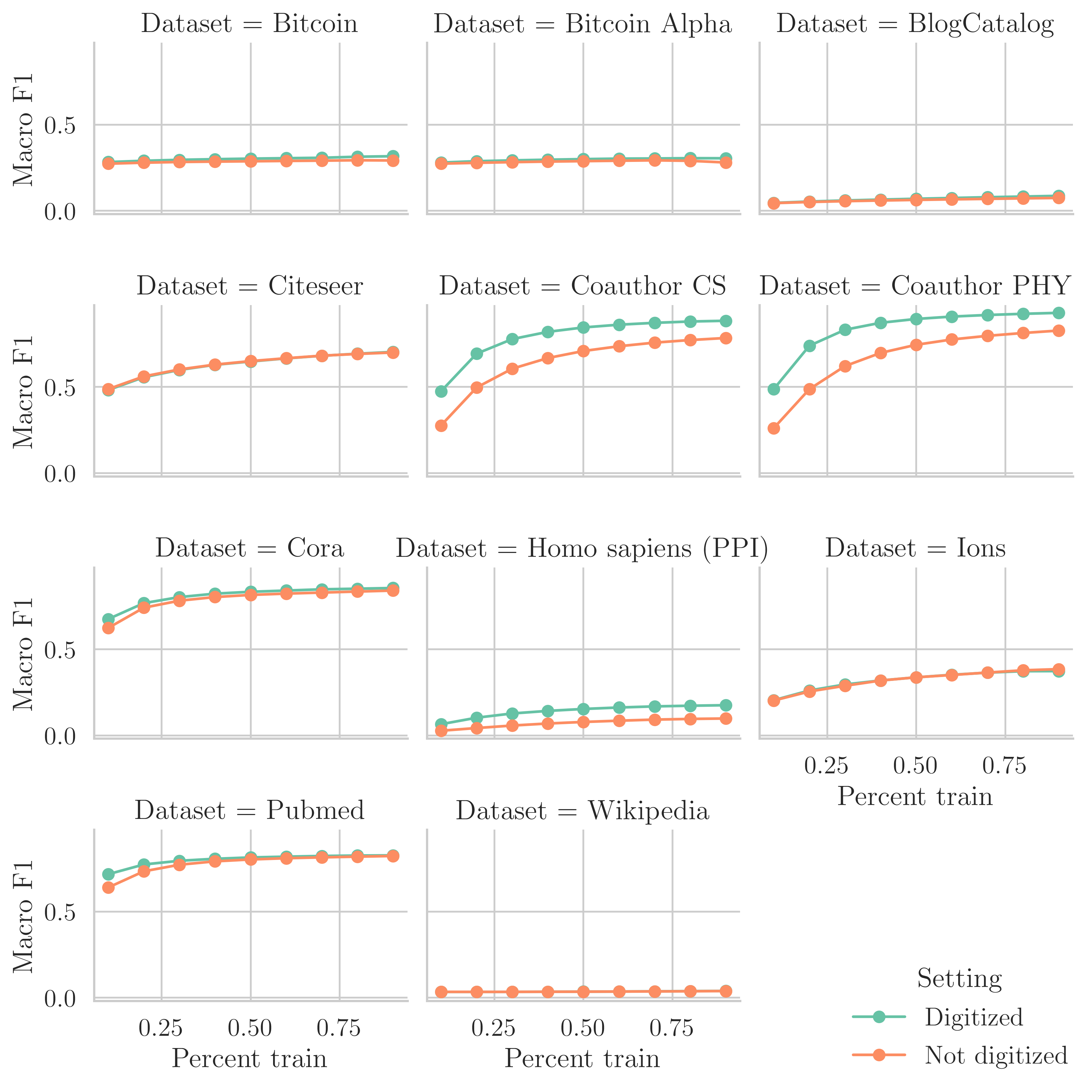}
  \caption{Micro F1 plots comparing digitized and non-digitized embedding.}
  \label{fig:macrodigitized}
\end{figure}

Fig.~\ref{fig:microdigitized} and~\ref{fig:macrodigitized} show how results performance is affected if embeddings are digitized as described in Section~\ref{sec:dim}. We can see that digitized embeddings usually perform similarly and even outperform non-digitized embeddings on both Co-authorship datasets, Homo sapiens (PPI) dataset, and Pubmed dataset.

\section{Average rank diagrams of mean classification results}
\label{sec:app-rank}

\begin{figure*}[t!]
  \centering
  \includegraphics[width = \linewidth]{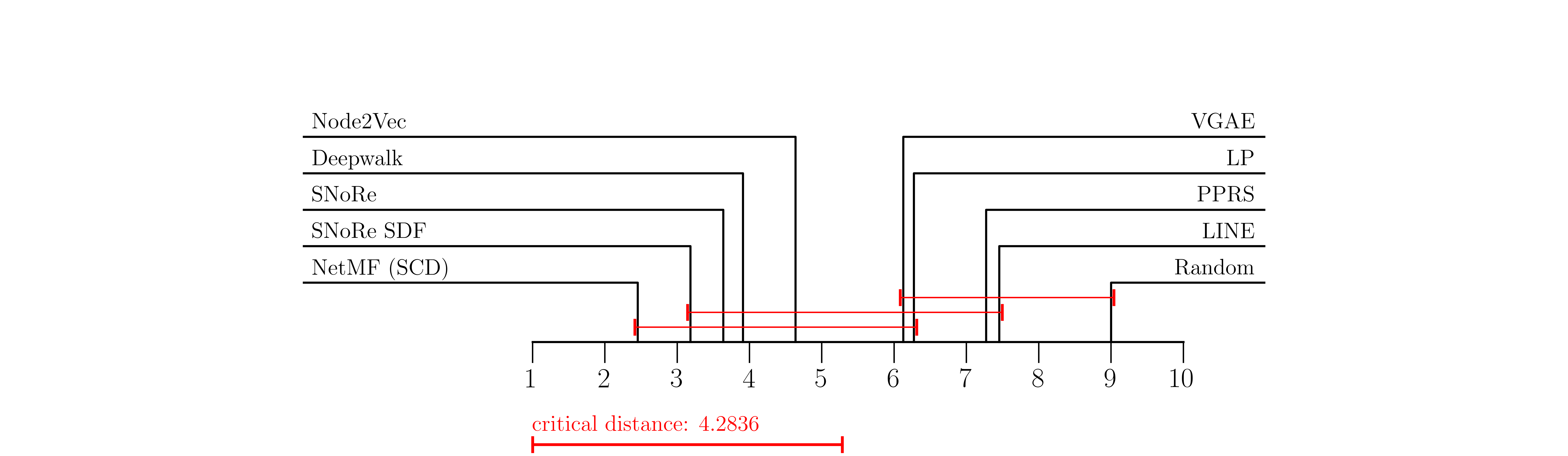}
  \caption{Micro F1 average rank diagram where average performance across all training percentages is chosen.}
  \label{fig:cdmicromean}
\end{figure*}

\begin{figure*}[t!]
  \centering
  \includegraphics[width = \linewidth]{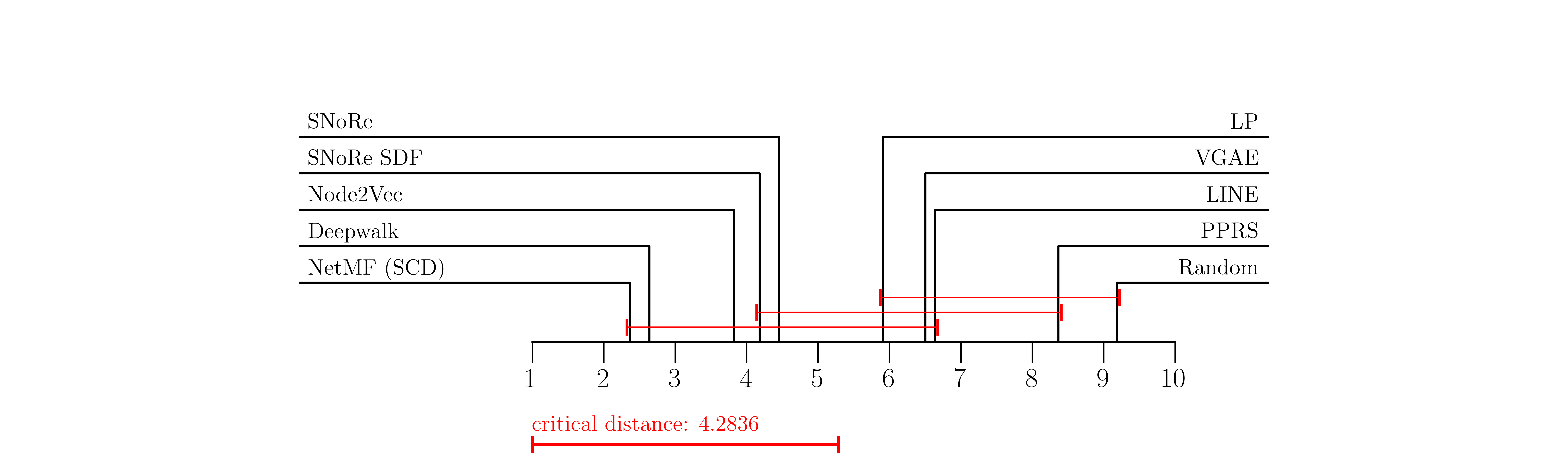}
  \caption{Macro F1 average rank diagram where average performance across all training percentages is chosen.}
  \label{fig:cdmacromean}
\end{figure*}

Fig.~\ref{fig:cdmicromean} and~\ref{fig:cdmacromean} show the average rank diagrams when we average classification results of every training set size. Here we see that SNoRe (SDF) and SNoRe achieve similar ranks in both micro and macro F1. We can see that here SNoRe (SDF) achieves worse results than on Fig.~\ref{fig:cdmicromax} and~\ref{fig:cdmacromax}. The reason behind this can be seen on Coauthor CS and Coauthor PHY datasets in Fig.~\ref{fig:micro}, where the algorithm performs poorly compared to others when a small amount of training data is used.

\section{Parameter study macro plots}
\label{sec:app-macro}

\begin{figure}[t!]
  \centering
  \includegraphics[width = \linewidth] {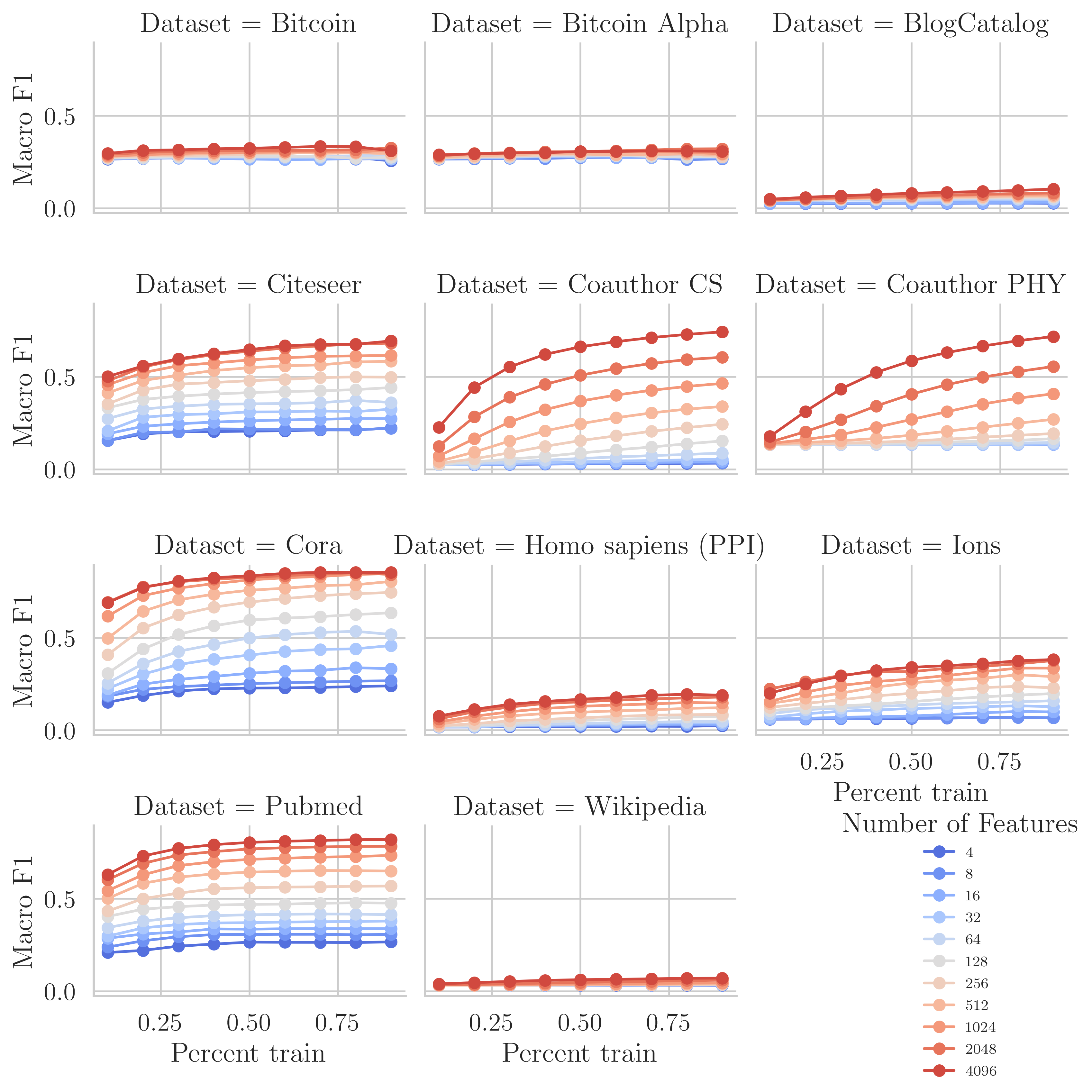}
  \caption{Macro F1 plots for different number of features (pivot nodes).}
  \label{fig:macrofeat}
\end{figure}

\begin{figure}[t!]
  \centering
  \includegraphics[width = \linewidth] {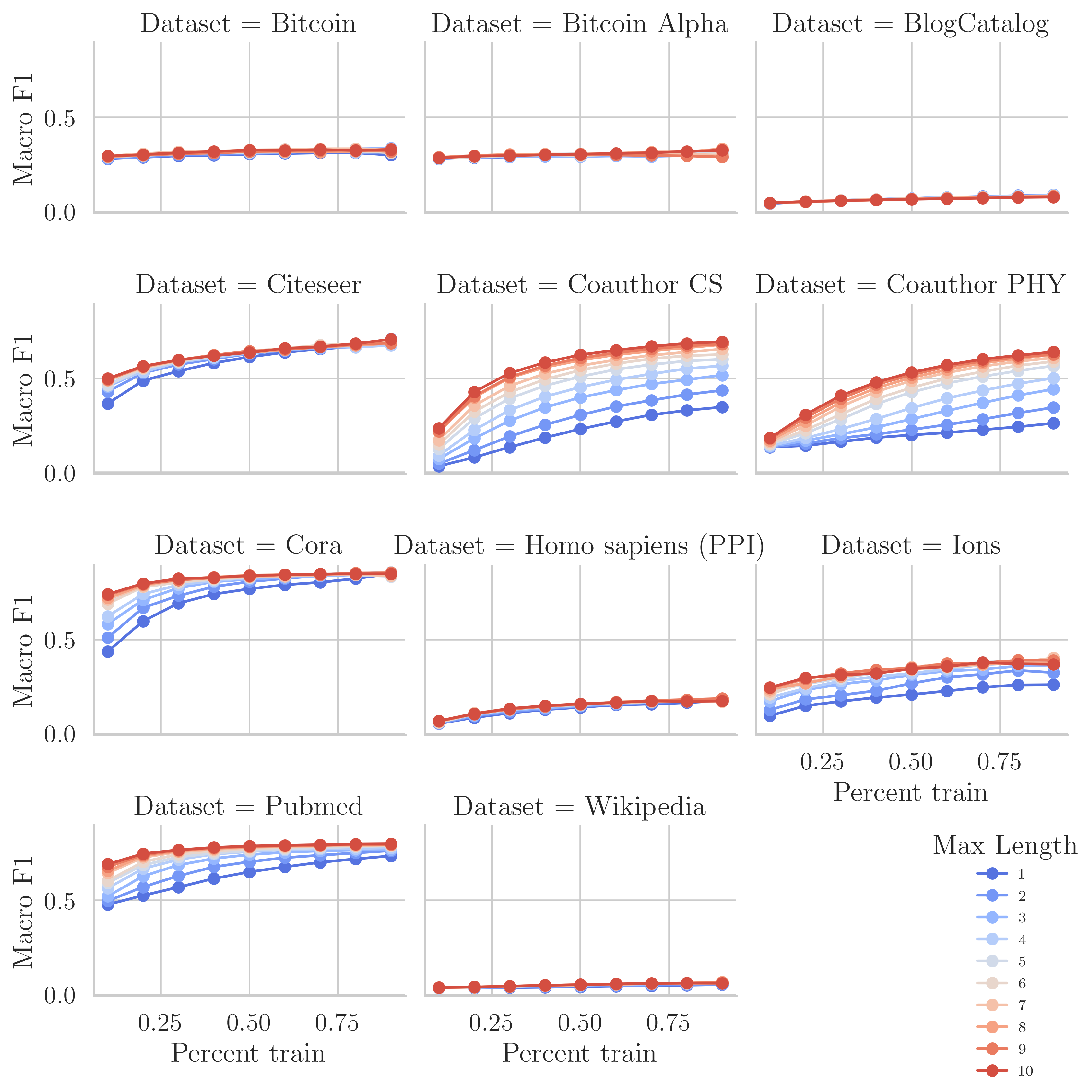}
  \caption{Macro F1 plots showing the effect of maximum walk length parameter.}
  \label{fig:macrolength}
\end{figure}

\begin{figure}[t!]
  \centering
  \includegraphics[width = \linewidth] {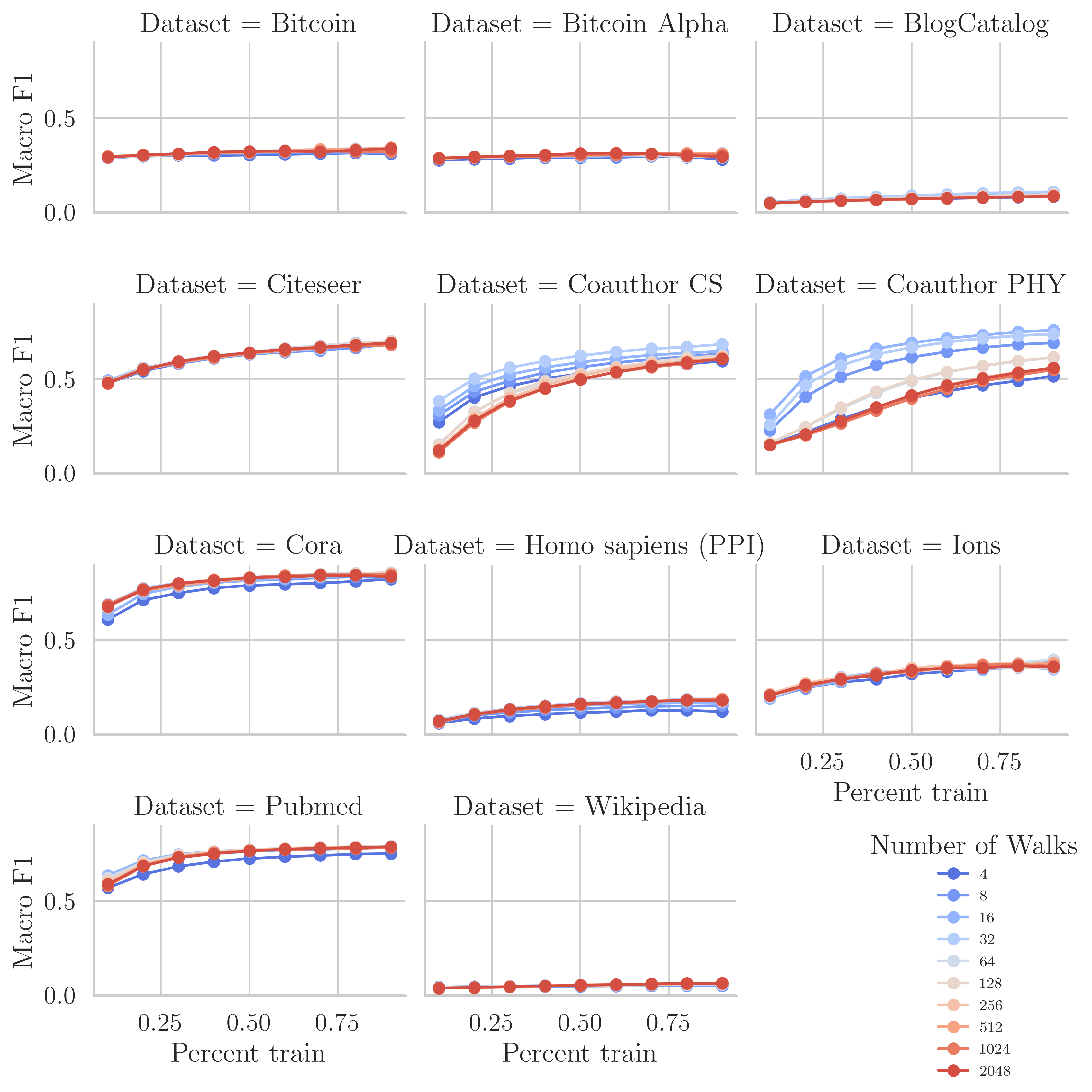}
  \caption{Macro F1 plots showing the effect of different number of random walks.}
  \label{fig:macrowalks}
\end{figure}

Fig.~\ref{fig:macrofeat},~\ref{fig:macrolength}, and~\ref{fig:macrowalks} show the macro F1 plots for different parameter settings. We can see that the overall parameter number of features (pivot nodes) impacts classification the most. On the other hand, the number of walks parameter only affects the macro F1 score on datasets Coauthor CS and Coauthor PHY. Datasets Bitcoin, Bitcoin Alpha, BlogCatalog, and Wikipedia have the same score no matter which parameters we use.

\section{Additional execution time plots}
\label{sec:app-time}
In this section, we show how different distance metrics and parameters affect execution time.

\subsection{Distance metrics}
\begin{figure}[h!]
  \centering
  \includegraphics[width = \linewidth]{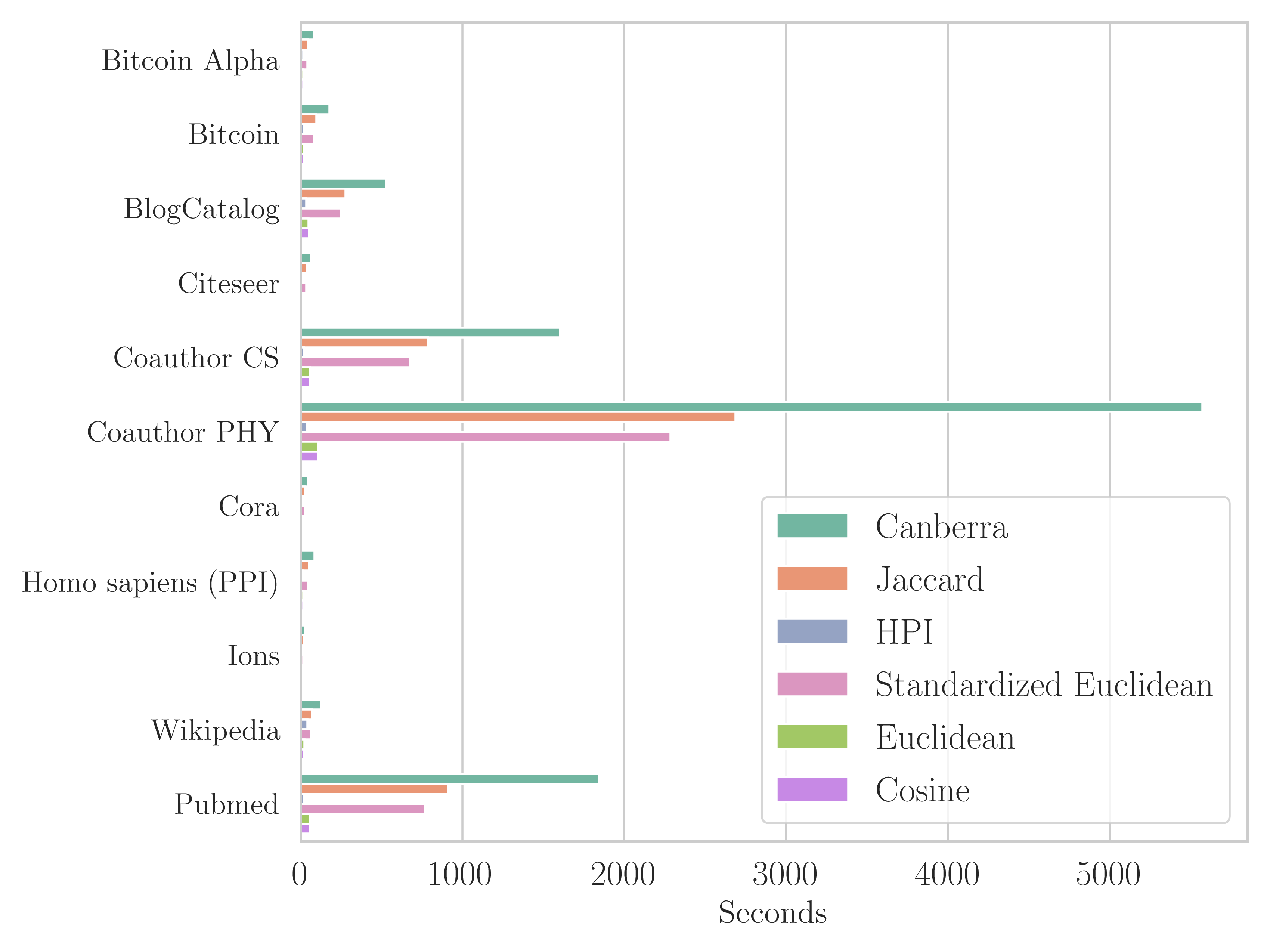}
  \caption{Time plot for different metrics.}
  \label{fig:timemetric}
\end{figure}

We present the effect of different distance metrics in Fig.~\ref{fig:timemetric}. We use sparse matrices for storing random walk hashes, which makes the implementation of similarity calculation crucial to obtain good performance. This can be seen in the figure, where Euclidean distance, HPI, and cosine similarity need significantly less time than other distance metrics that are not optimized for sparse matrices. Here we would like to highlight that the HPI distance metric we implemented even less time to execute than the implementation of sparse euclidean distance and cosine similarity from the highly optimized Python library scikit learn.

\subsection{Number of pivot nodes}
\begin{figure}[t!]
  \centering
  \includegraphics[width = \linewidth]{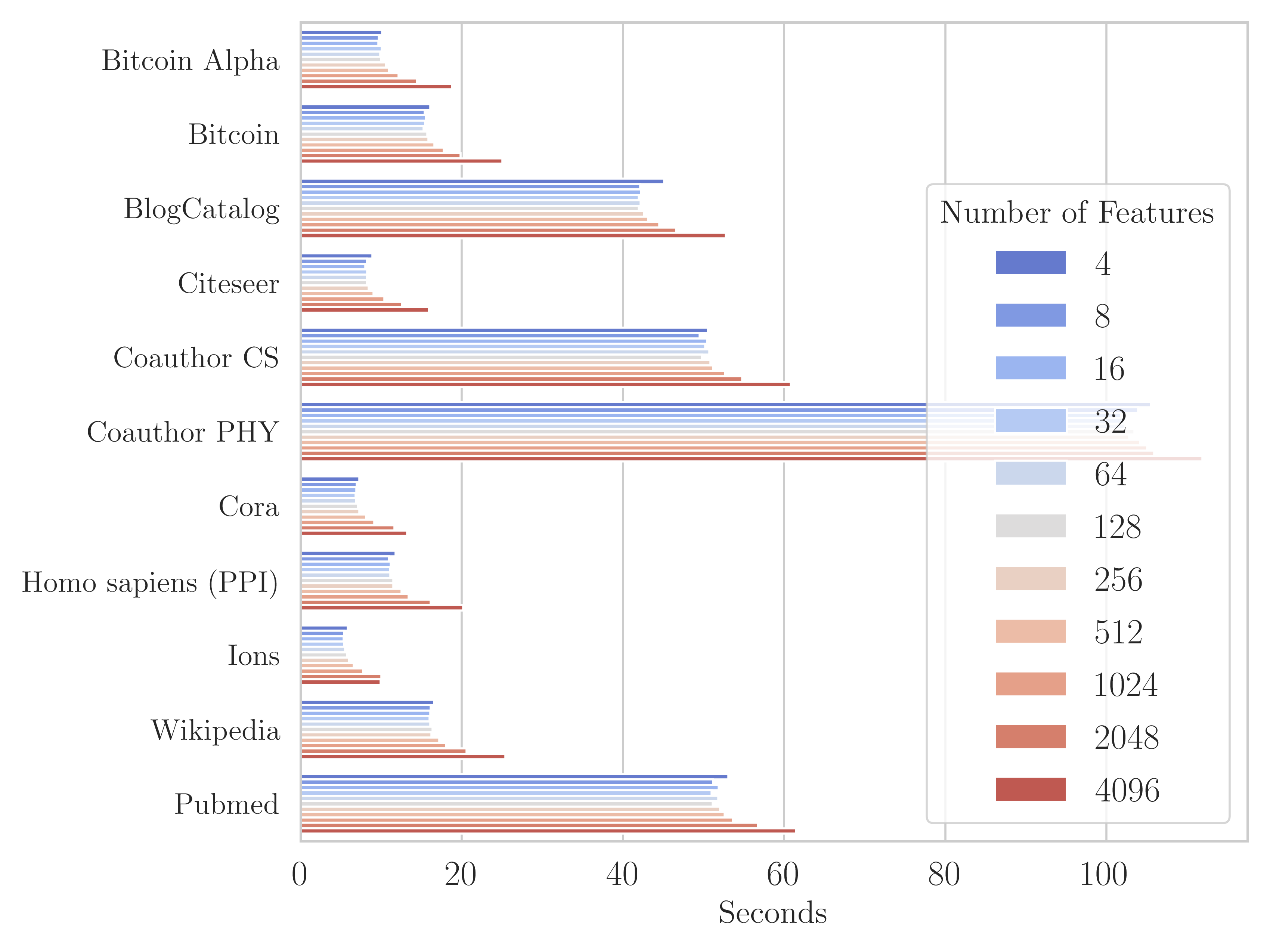}
  \caption{Time plot of different number of features (pivot nodes).}
  \label{fig:timenf}
\end{figure}

Fig.~\ref{fig:timenf} shows how the parameter number of pivot nodes affects execution time. From the figure, we see that the difference between different values of the parameter is not significant and that the impact of the number of nodes is far greater. A small impact of the different number of pivot nodes on execution time gives us further reason to use SNoRe (SDF) since execution time stays similar.

\subsection{Maximum walk length}
\begin{figure}[t!]
  \centering
  \includegraphics[width = \linewidth]{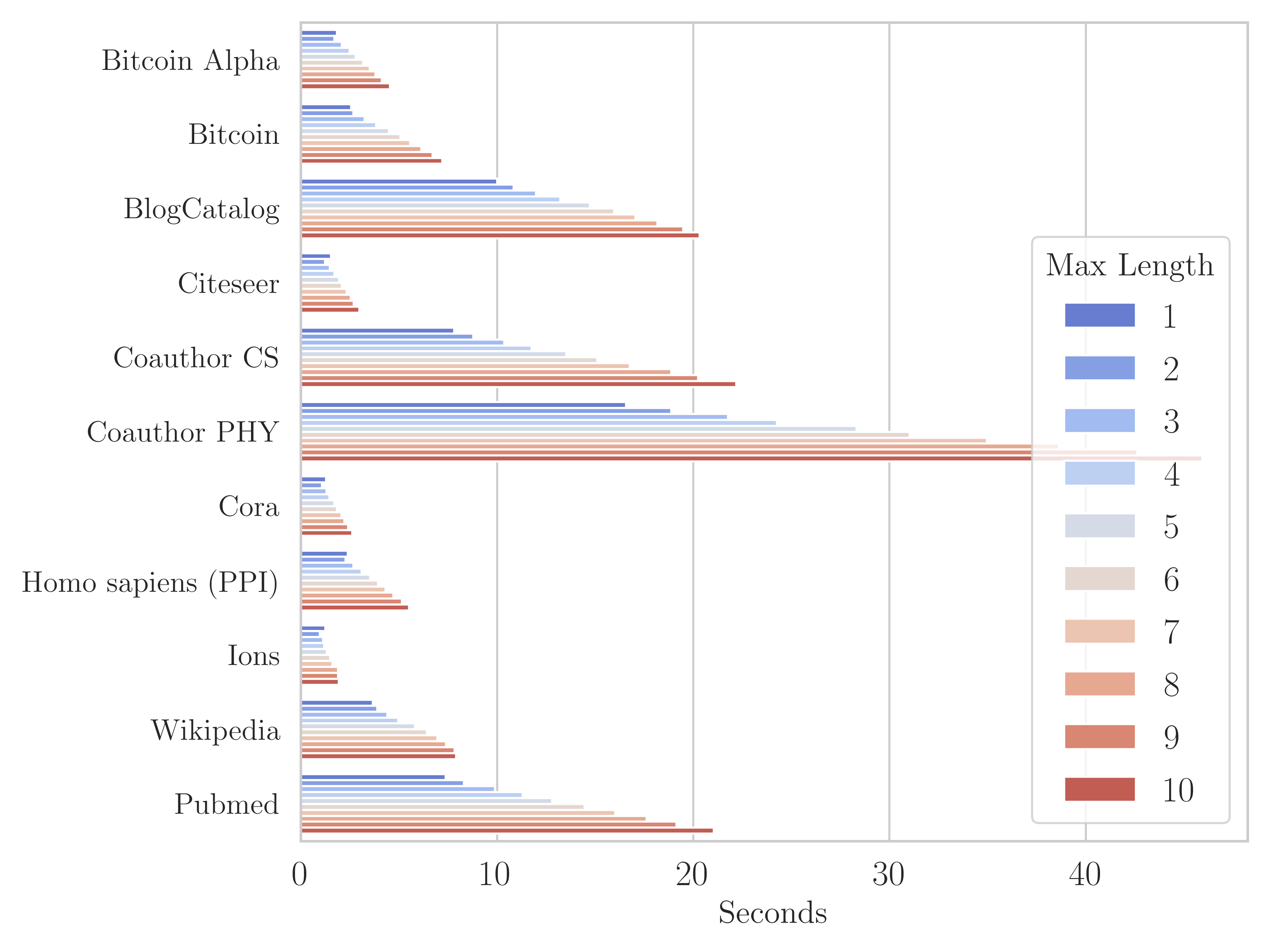}
  \caption{Time plot of different values of maximum walk length parameter.}
  \label{fig:timelength}
\end{figure}

The execution time is also affected by the maximum walk length parameter. The execution time between different values of the parameter can be seen in Fig.~\ref{fig:timelength}. The execution time rises linearly on all datasets. Linear rise is expected since we sample walks from the uniform distribution, making $\overline{s}=\frac{\textrm{walk length + 1}}{2}$. As seen in Section~\ref{sec:prop-time} $\overline{s}$ affects time linearly.

\subsection{Number of walks per node}
\begin{figure}[t!]
  \centering
  \includegraphics[width = \linewidth]{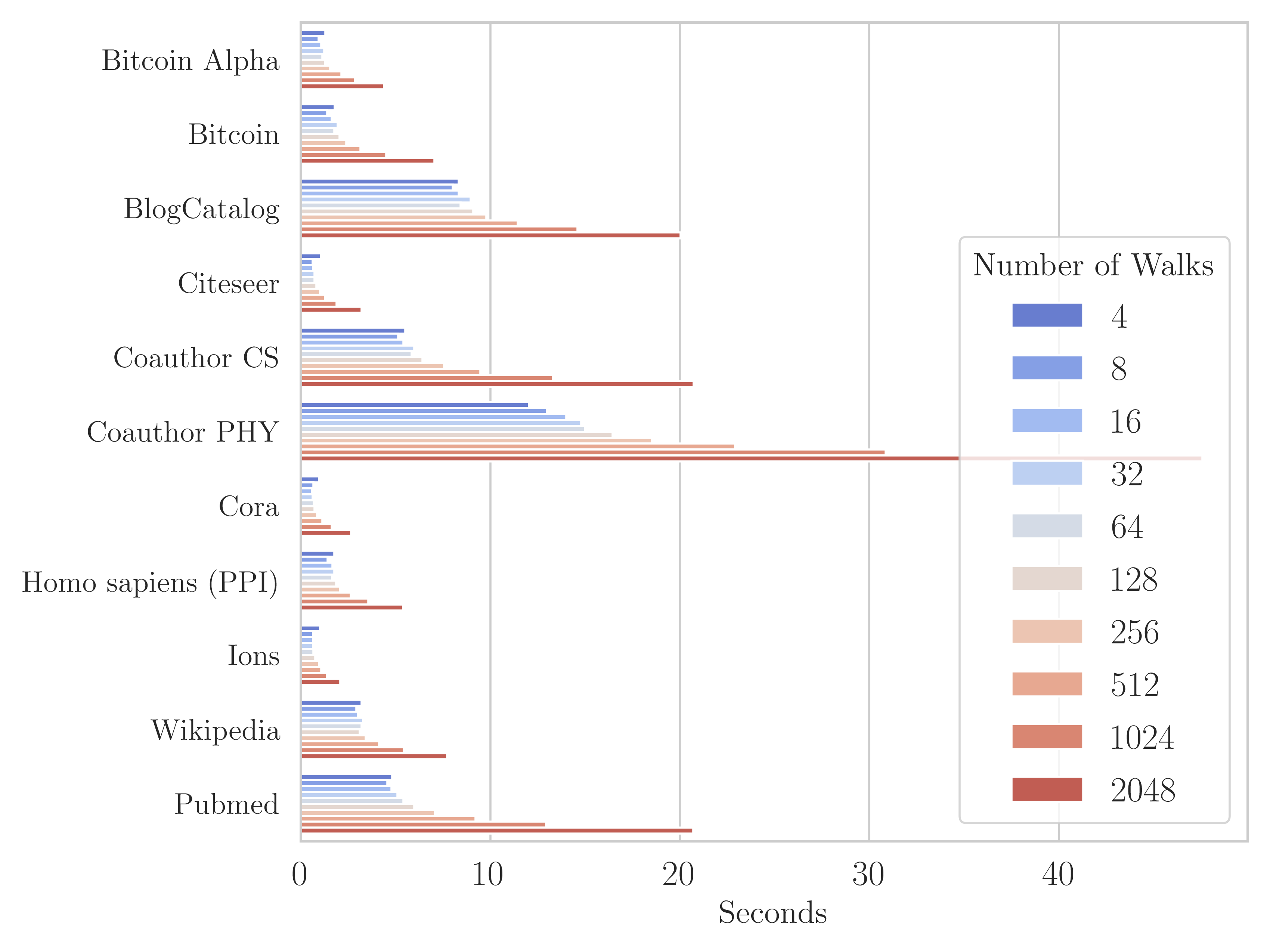}
  \caption{Time plot for different number of random walks.}
  \label{fig:timewalks}
\end{figure}

The last parameter we show is the number of random walks per node. The effects of different values of this parameter on execution time can be seen in Fig.~\ref{fig:timewalks}. We see that execution time grows linearly with the number of walks per node. Linear growth is expected and further backs the claim made in Section~\ref{sec:prop-time}.

\section{Classification explanation with SHAP}
\label{sec:app-shap}

\begin{figure*}[h!]
  \centering
  \includegraphics[width = \linewidth]{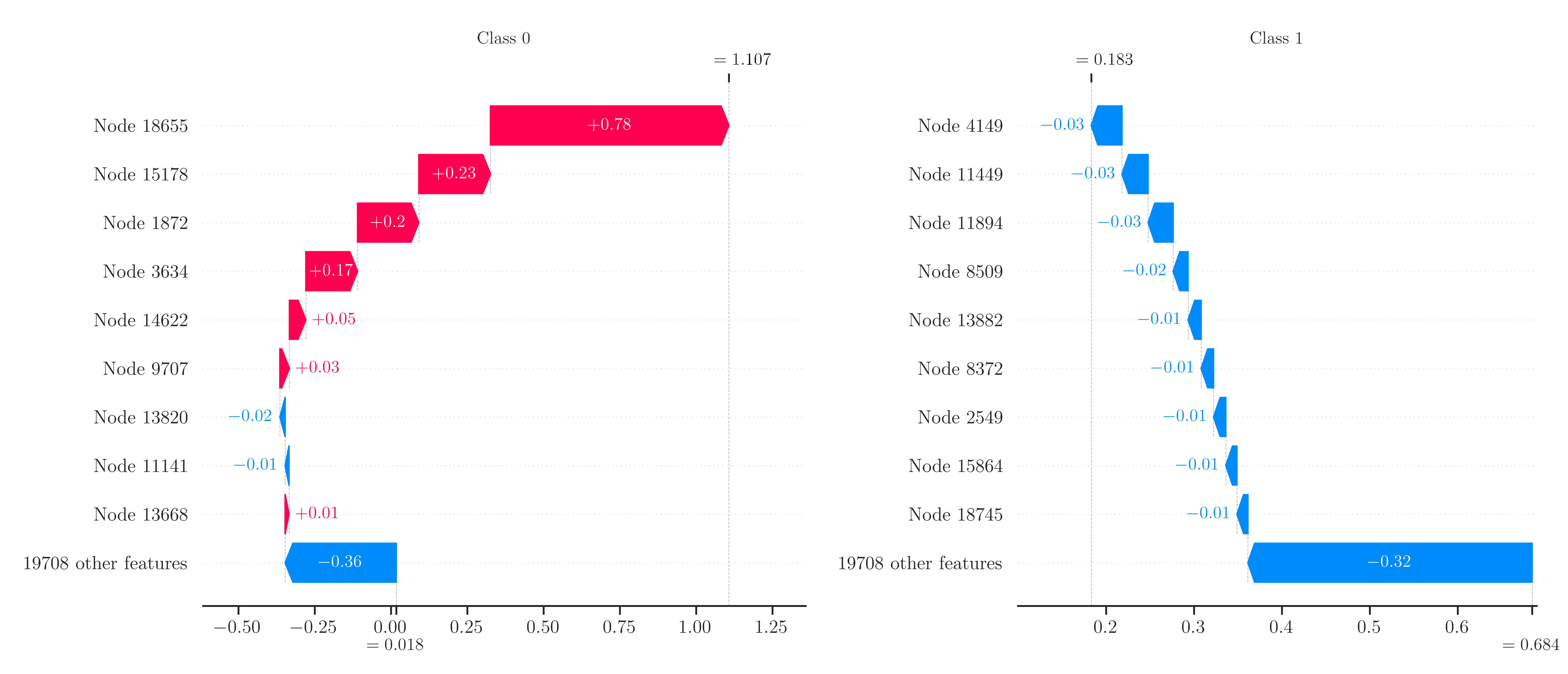}
  \caption{Waterfall explanation for classification of a node.}
  \label{fig:watefall}
\end{figure*}

Fig.~\ref{fig:watefall} shows how we can interpret the classification of an instance using SHAP. In the figure, we see two examples, one where class (label) 0 is assigned, and one where class (label) 1 is not assigned. The classification for class 0 starts at the expected value of 0.018. Then the value of features that represent nodes 13820, 11141, and some others lower this value for around 0.39. Values of features that represent nodes 13668, 9707, ..., 15178, and 18655 raise the value by 1.47 to the final value 1.107. Since the final score is high the class (label) 0 is assigned to the instance. We see that the value of feature that represents node 18655 has the biggest impact and that class (label) 0 is mostly assigned because of the high (or less likely low) similarity between the neighborhood of node 18655 and the observed node (instance).

Class (label) 1 is not assigned to this instance. We can see that the classification starts at the expected value of 0.684 for class 1 and is only lowered. All features lower the score of the classification from 0.501 to the final 0.183. Since this score is low, the class (label) 1 is not assigned to this node. We can see that the prediction is lowered the most by features that represent nodes 4149, 11449, and 11894.

\end{document}

%% file: mezna.t1.tex
\resizebox{\textwidth}{!}{
\begin{tabular}{lllllllllll}
\toprule
setting &           Random &               LP &             VGAE &             PPRS &             LINE &         Node2Vec &         Deepwalk &      NetMF (SCD) &            SNoRe & SNoRe SDF\\
dataset            &                  &                  &                  &                  &                  &                  &                  &                  &                  \\
\midrule
Bitcoin            &   0.670 ($\pm$0.011) &  0.701 ($\pm$0.003) &    0.700 ($\pm$0.004) &  0.692 ($\pm$0.003) &  0.662 ($\pm$0.008) &  0.687 ($\pm$0.012) &  0.696 ($\pm$0.016) &  0.703 ($\pm$0.009) &\bfseries  0.716 ($\pm$0.007) & 0.709 ($\pm$ 0.007)\\
Bitcoin Alpha      &  0.676 ($\pm$0.013) &  0.701 ($\pm$0.002) &  0.694 ($\pm$0.005) &  0.699 ($\pm$0.003) &  0.665 ($\pm$0.005) &  0.678 ($\pm$0.011) &  0.683 ($\pm$0.017) &   0.694 ($\pm$0.010) &\bfseries  0.709 ($\pm$0.005) &  0.703 ($\pm$ 0.003)\\
BlogCatalog        &  0.139 ($\pm$0.014) &     0.070 ($\pm$0.000) &              OOM &  0.169 ($\pm$0.002) &   0.289 ($\pm$0.030) &  0.373 ($\pm$0.012) &  0.385 ($\pm$0.022) &\bfseries   0.420 ($\pm$0.017) &   0.230 ($\pm$0.021) &  0.226 ($\pm$ 0.019) \\
Citeseer           &  0.187 ($\pm$0.007) &  0.657 ($\pm$0.062) &  0.405 ($\pm$0.013) &  0.334 ($\pm$0.027) &  0.299 ($\pm$0.017) &  0.583 ($\pm$0.024) &  0.578 ($\pm$0.027) &   0.590 ($\pm$0.019) &\bfseries  0.666 ($\pm$0.063) &  0.664 ($\pm$ 0.066)\\
Coauthor CS        &  0.215 ($\pm$0.014) &    0.125 ($\pm$0.000) &  0.773 ($\pm$0.013) &  0.227 ($\pm$0.002) &  0.677 ($\pm$0.026) &  0.878 ($\pm$0.008) &\bfseries  0.883 ($\pm$0.009) &\bfseries  0.883 ($\pm$0.006) &  0.585 ($\pm$0.131) &  0.854 ($\pm$ 0.081)\\
Coauthor PHY       &  0.505 ($\pm$0.001) &    0.333 ($\pm$0.000) &              OOM &  0.505 ($\pm$0.002) &   0.754 ($\pm$0.010) &  0.931 ($\pm$0.003) &\bfseries  0.935 ($\pm$0.003) &\bfseries  0.935 ($\pm$0.002) &  0.605 ($\pm$0.061) &  0.887 ($\pm$ 0.082)\\
Cora               &  0.246 ($\pm$0.021) &\bfseries  0.834 ($\pm$0.038) &  0.645 ($\pm$0.023) &  0.445 ($\pm$0.041) &  0.432 ($\pm$0.028) &  0.809 ($\pm$0.019) &  0.817 ($\pm$0.026) &  0.822 ($\pm$0.022) &  0.822 ($\pm$0.052) &  0.826 ($\pm$ 0.051)\\
Homo sapiens (PPI) &  0.061 ($\pm$0.003) &    0.066 ($\pm$0.000) &   0.167 ($\pm$0.010) &  0.113 ($\pm$0.013) &  0.143 ($\pm$0.014) &  0.205 ($\pm$0.019) &  0.205 ($\pm$0.023) &\bfseries  0.227 ($\pm$0.022) &  0.207 ($\pm$0.041) &  0.210 ($\pm$ 0.038)\\
Ions               &   0.383 ($\pm$0.020) &  0.691 ($\pm$0.051) &  0.569 ($\pm$0.018) &  0.529 ($\pm$0.026) &   0.640 ($\pm$0.032) &  0.661 ($\pm$0.029) &  0.685 ($\pm$ 0.027) &  0.706 ($\pm$0.031) &\bfseries  0.712 ($\pm$0.045) &  0.708 ($\pm$ 0.047)\\
Pubmed             &  0.395 ($\pm$0.003) &  0.399 ($\pm$0.001) &  0.676 ($\pm$0.015) &  0.398 ($\pm$0.001) &  0.611 ($\pm$0.013) &  0.804 ($\pm$0.005) &  0.806 ($\pm$0.005) &  0.813 ($\pm$0.005) &  0.783 ($\pm$0.033) &\bfseries  0.821 ($\pm$ 0.024)\\
Wikipedia          &  0.394 ($\pm$0.014) &    0.068 ($\pm$0.000) &              OOM &  0.441 ($\pm$0.011) &  0.382 ($\pm$0.015) &\bfseries  0.505 ($\pm$0.015) &  0.465 ($\pm$0.023) &  0.501 ($\pm$0.016) &  0.427 ($\pm$0.014) &  0.404 ($\pm$ 0.002)\\
\bottomrule
\end{tabular}
}

%% file: mezna.t2.tex
\resizebox{\textwidth}{!}{
\begin{tabular}{lllllllllll}
\toprule
setting &           Random &               LP &             VGAE &             PPRS &             LINE &         Node2Vec &         Deepwalk &      NetMF (SCD) &            SNoRe & SNoRe SDF\\
dataset            &                  &                  &                  &                  &                  &                  &                  &                  &                  \\
\midrule
Bitcoin            &  0.269 ($\pm$0.004) &  0.287 ($\pm$0.002) &  0.304 ($\pm$0.006) &  0.277 ($\pm$0.006) &  0.293 ($\pm$0.008) &  0.315 ($\pm$0.012) &\bfseries  0.318 ($\pm$0.012) &  0.312 ($\pm$0.009) &  0.314 ($\pm$0.013) & 0.293 ($\pm$ 0.009) \\
Bitcoin Alpha      &   0.270 ($\pm$0.006) &  0.277 ($\pm$0.004) &  0.288 ($\pm$0.005) &  0.282 ($\pm$0.007) &  0.282 ($\pm$0.006) &  0.296 ($\pm$0.009) &  0.299 ($\pm$0.011) &  0.299 ($\pm$0.009) &\bfseries  0.303 ($\pm$0.009) & 0.283 ($\pm$ 0.005) \\
BlogCatalog        &  0.037 ($\pm$0.004) &    0.068 ($\pm$0.000) &              OOM &  0.027 ($\pm$0.001) &  0.169 ($\pm$0.022) &  0.206 ($\pm$0.017) &  0.243 ($\pm$0.025) &\bfseries  0.271 ($\pm$0.022) &  0.067 ($\pm$0.012) & 0.065 ($\pm$ 0.011) \\
Citeseer           &  0.157 ($\pm$0.006) &    0.620 ($\pm$0.060) &  0.344 ($\pm$0.011) &   0.270 ($\pm$0.038) &  0.255 ($\pm$0.018) &  0.532 ($\pm$0.023) &  0.532 ($\pm$0.025) &   0.540 ($\pm$0.019) &  0.621 ($\pm$0.066) & \bfseries0.623 ($\pm$ 0.067)\\
Coauthor CS        &  0.032 ($\pm$0.009) &     0.120 ($\pm$0.000) &  0.662 ($\pm$0.021) &  0.026 ($\pm$0.002) &  0.622 ($\pm$0.036) &  0.849 ($\pm$0.011) &\bfseries  0.855 ($\pm$0.013) &  0.853 ($\pm$0.008) &  0.451 ($\pm$0.155) &  0.803 ($\pm$ 0.109)\\
Coauthor PHY       &    0.134 ($\pm$0.000) &    0.309 ($\pm$0.000) &              OOM &    0.134 ($\pm$0.000) &  0.675 ($\pm$0.013) &  0.908 ($\pm$0.003) &  0.912 ($\pm$0.004) &\bfseries  0.914 ($\pm$0.003) &  0.381 ($\pm$0.141) &  0.853 ($\pm$ 0.118)\\
Cora               &  0.117 ($\pm$0.014) &\bfseries  0.825 ($\pm$0.038) &  0.616 ($\pm$0.029) &  0.351 ($\pm$0.073) &  0.366 ($\pm$0.043) &  0.799 ($\pm$0.021) &  0.808 ($\pm$0.028) &  0.812 ($\pm$0.024) &  0.811 ($\pm$0.054) &  0.815 ($\pm$ 0.054)\\
Homo sapiens (PPI) &  0.046 ($\pm$0.002) &    0.066 ($\pm$0.000) &   0.104 ($\pm$0.010) &  0.065 ($\pm$0.015) &  0.121 ($\pm$0.015) &  0.173 ($\pm$0.019) &  0.174 ($\pm$0.022) &\bfseries  0.189 ($\pm$0.022) &  0.142 ($\pm$0.037) &  0.156 ($\pm$ 0.037)\\
Ions               &  0.076 ($\pm$0.005) &\bfseries  0.333 ($\pm$0.031) &  0.163 ($\pm$0.013) &  0.176 ($\pm$0.036) &  0.288 ($\pm$0.031) &   0.299 ($\pm$0.030) &   0.321 ($\pm$ 0.026) &  0.309 ($\pm$0.029) &  0.319 ($\pm$0.053) &  0.312 ($\pm$ 0.052)\\
Pubmed             &  0.295 ($\pm$0.004) &     0.190 ($\pm$0.000) &  0.641 ($\pm$0.018) &     0.190 ($\pm$0.000) &  0.567 ($\pm$0.011) &   0.790 ($\pm$0.005) &  0.792 ($\pm$0.006) &    0.800 ($\pm$0.006) &  0.742 ($\pm$0.059) &\bfseries  0.805 ($\pm$ 0.032)\\
Wikipedia          &  0.041 ($\pm$0.003) &    0.059 ($\pm$0.000) &              OOM &   0.080 ($\pm$0.011) &  0.058 ($\pm$0.004) &   0.099 ($\pm$0.010) &  0.087 ($\pm$0.008) &\bfseries  0.103 ($\pm$0.008) &   0.050 ($\pm$0.009) &  0.034 ($\pm$ 0.001)\\
\bottomrule
\end{tabular}
}